\documentclass[draft,final]{vutinfth} % Remove option 'final' to obtain debug information.
\usepackage{lmodern}        % Use an extension of the original Computer Modern font to minimize the use of bitmapped letters.
\usepackage[T1]{fontenc}    % Determines font encoding of the output. Font packages have to be included before this line.
\usepackage[utf8]{inputenc} % Determines encoding of the input. All input files have to use UTF8 encoding.

\usepackage{amsmath}    % Extended typesetting of mathematical expression.
\usepackage{amssymb}    % Provides a multitude of mathematical symbols.
\usepackage{mathtools}  % Further extensions of mathematical typesetting.
\usepackage{microtype}  % Small-scale typographic enhancements.
\usepackage[inline]{enumitem} % User control over the layout of lists (itemize, enumerate, description).
\usepackage{multirow}   % Allows table elements to span several rows.
\usepackage{booktabs}   % Improves the typesettings of tables.
\usepackage{subcaption} % Allows the use of subfigures and enables their referencing.
\usepackage[ruled,linesnumbered,algochapter]{algorithm2e} % Enables the writing of pseudo code.
\usepackage[usenames,dvipsnames,table]{xcolor} % Allows the definition and use of colors. This package has to be included before tikz.
\usepackage{nag}       % Issues warnings when best practices in writing LaTeX documents are violated.
\usepackage{todonotes} % Provides tooltip-like todo notes.
\usepackage{hyperref}  % Enables cross linking in the electronic document version. This package has to be included second to last.
\usepackage[acronym,toc]{glossaries} % Enables the generation of glossaries and lists fo acronyms. This package has to be included last.

\newcommand{\authorname}{Jovan Jeromela} % The author name without titles.
\newcommand{\thesistitle}{Semi-Abstract Value-Based Argumentation Framework} % The title of the thesis. The English version should be used, if it exists.

\newcommand{\change}[1]{#1}
\newcommand{\vthree}[1]{#1}
\hypersetup{
    pdfpagelayout   = TwoPageRight,           % How the document is shown in PDF viewers (optional).
    linkbordercolor = {Melon},                % The color of the borders of boxes around crosslinks (optional).
    pdfauthor       = {\authorname},          % The author's name in the document properties (optional).
    pdftitle        = {\thesistitle},         % The document's title in the document properties (optional).
    pdfsubject      = {Subject},              % The document's subject in the document properties (optional).
    pdfkeywords     = {a, list, of, keywords} % The document's keywords in the document properties (optional).
}

\setpnumwidth{2.5em}        % Avoid overfull hboxes in the table of contents (see memoir manual).
\setsecnumdepth{subsection} % Enumerate subsections.

\nonzeroparskip             % Create space between paragraphs (optional).
\makeindex      % Use an optional index.
\makeglossaries % Use an optional glossary.
\setauthor{ }{\authorname}{ }{male}
\setadvisor{Ao.Univ.Prof. Dipl.-Ing. Dr.techn.}{Christian Fermüller}{ }{male}

\setfirstreviewer{Pretitle}{Forename Surname}{Posttitle}{male}
\setsecondreviewer{Pretitle}{Forename Surname}{Posttitle}{male}

\setsecondadvisor{Pretitle}{Forename Surname}{Posttitle}{male} % Comment to remove.

\setaddress{Lorenz-Müller-Gasse 1A/5283, A-1200 Wien, Österreich}
\setregnumber{01528091}
\setdate{7}{11}{2019} % Set date with 3 arguments: {day}{month}{year}.
\settitle{\thesistitle}{Semi-Abstract Value-Based Argumentation Framework} % Sets English and German version of the title (both can be English or German). If your title contains commas, enclose it with additional curvy brackets (i.e., {{your title}}) or define it as a macro as done with \thesistitle.
\setthesis{bachelor}
\setcurriculum{Software \& Information Engineering}{Software \& Information Engineering} % Sets the English and German name of the curriculum.

\setfirstreviewerdata{Affiliation, Country}
\setsecondreviewerdata{Affiliation, Country}

\usepackage{longtable} % https://tex.stackexchange.com/questions/26462/make-a-table-span-multiple-pages
\allowdisplaybreaks

\begin{document}

\frontmatter % Switches to roman numbering.
% The structure of the thesis has to conform to
%  http://www.informatik.tuwien.ac.at/dekanat

\addtitlepage{naustrian} % German title page (not for dissertations at the PhD School).
\addtitlepage{english} % English title page.
\addstatementpage

\begin{danksagung*}
Zuallererst möchte ich mich bei meinem Mentor Prof. Dr. Christian Fermüller für seine wertvolle Betreuung während des gesamten Prozesses der Abfassung dieser Arbeit, für seine stets rasche Beantwortung meiner Fragen und für das Verständnis für Hindernisse, die die für die Fertigstellung dieser Arbeit erforderliche Zeit verlängert haben, bedanken. 

Ich möchte auch meiner gesamten Familie -- insbesondere meinen Eltern und Großeltern -- von ganzem Herzen für die finanzielle und emotionale Unterstützung danken, ohne die mein Studienabschluss nicht möglich wäre.

% Ich möchte mich auch bei Jasmina Gajčin bedanken, die mir geholfen hat, die gesamte Arbeit (mehrmals) zu korrigieren.

%Abschließend möchte ich meiner gesamten Familie -- insbesondere meinen Eltern und Großeltern -- von ganzem Herzen für die finanzielle und emotionale Unterstützung danken, ohne die mein Studienabschluss nicht möglich wäre.

\end{danksagung*}

\begin{acknowledgements*}
Foremost, I want to thank my supervisor prof. dr Christian Fermüller for his valuable guidance during the entire process of writing this thesis, for his always-quick responses to my questions, and for the understanding regarding obstacles which prolonged the time requisite for the completion of this work.

I also want to thank my entire family -- especially my parents and grandparents -- from the bottom of my heart for the encouragement and support, both financial and emotional, without which none of this would have been possible.

% I would also like to thank Jasmina Gajčin for helping me proofread the entire thesis (multiple times).

% Lastly, I want to thank my entire family -- especially my parents and grandparents -- from the bottom of my heart for the encouragement and support, both financial and emotional, without which none of this would have been possible.
\end{acknowledgements*}

\begin{kurzfassung}

Phan Minh Dung hat in seiner grundlegenden Arbeit \cite{dung95} das \emph{abstract Argumentation Framework} vorgestellt. Dieses Framework modelliert Argumentation mittels geordneten Graphen, wo die strukturlosen Argumente als Knoten und die Angriffe zwischen Argumenten als Kanten dargestellt sind. In den folgen Jahren wurden zahlreiche Erweiterungen dieses Frameworks eingebracht. Diese Erweiterungen fügen typischerweise manche Gestaltung zu den Argumenten hinzu. Diese Diplomarbeit präsentiert zwei solche Erweiterungen -- \emph{value-based Argumentation \change{Framework}} von Trevor Bench-Capon \cite{capon02} und \emph{semi-abstract Argumentation \change{Framework}} von Esther Anna Corsi und Christian Ferm\"uller \cite{corfer17lori}. Die erstere verbindet einzelne Argumente mit geordneten Werten, damit man zwischen subjektiv und objektiv akzeptablen Argumenten differenzieren kann. Die letztere fängt von Argumenten, \change{deren \vthree{Behauptungen}} mit je einer aussagenlogischen Formel beschrieben sind. Dann wurde die Menge von neugegründete Angriffsprinzipien verwendet, um die impliziten Angriffe explizit zu machen und um eine Konsequenzrelation\change{,} die weder \change{die} Wahrheitswerte noch die gewöhnlichen Interpretationen braucht\change{,} definieren zu können.

Der Beitrag dieser Bachelorarbeit ist zweifältig. Erstens wurde das neue \emph{semi-abstract value-based Argumentation Framework} eingefürt. Dieses Framework bildet aussagenlogischen Formeln, die mit individuellen Argumenten assoziiert sind, zu geordneten Werten ab. Zweitens wurde ein komplexes moralisches Dilemma \change{durch das orignale und das value-based Frameworks präsentiert, um die Ausdrucksstärke dieser Formalismen} zu präsentieren.

\end{kurzfassung}

\begin{abstract}

In his seminal paper \cite{dung95}, Phan Minh Dung proposed \emph{abstract argumentation} framework, which models argumentation using directed graphs where structureless arguments are the nodes and attacks among the arguments are the edges. In the following years, many extensions of this framework were introduced. These extensions typically add a certain form of structure to the arguments. This thesis showcases two such extensions -- \emph{value-based argumentation \change{framework}} by Trevor Bench-Capon \cite{capon02} and \emph{semi-abstract argumentation \change{framework}} by Esther Anna Corsi and Christian Ferm\"uller \cite{corfer17lori}. The former introduces a mapping function that links individual arguments to \vthree{a} set of ordered values, \change{enabling a} distinction between objectively and subjectively acceptable arguments. The latter links \change{claims of} individual arguments to propositional formulae and then applies newly-introduced attack principles in order to make implicit attacks explicit and to enable \change{a definition of} a consequence relation \change{that relies} on neither the truth values nor the interpretations in the usual sense.

The contribution of this thesis is two-fold. Firstly, the new \emph{semi-abstract value-based argumentation framework} is introduced. This framework maps propositional formulae associated with individual arguments to a set of ordered values. Secondly, a complex moral dilemma is formulated using \change{the original and the value-based argumentation frameworks showcasing the expressivity of these formalisms}.

\end{abstract}

% Select the language of the thesis, e.g., english or naustrian.
\selectlanguage{english}

% Add a table of contents (toc).
\tableofcontents % Starred version, i.e., \tableofcontents*, removes the self-entry.

% Switch to arabic numbering and start the enumeration of chapters in the table of content.
\mainmatter

\chapter{Introduction}

\change{The authors of \cite{van2017argumentation} offer an overview of the history of the study of argumentation. The authors note that argumentation is a phenomenon which} is closely connected with differences in opinion and arises from the need to defend a certain standpoint and attack the opposing alternative. Although argumentation has been studied since Antiquity, during the 1970s the related research significantly intensified. This study was initially in opposition to the logical formalisms, as the expressivity necessary for adequately representing argumentation was lacking. This finally changed during the 1990s when progress in the fields of artificial intelligence resulted in the formation of formal argumentation frameworks.

 Nowadays, many systematic introductions to formal arguments \change{such as \cite{prakken2017historical}} start with Dung’s theory of abstract argumentation frameworks from \cite{dung95}, which can be thought of as graphs whose nodes are abstract (i.e. unstructured) \change{representations of} arguments and edges are the attack relations between \change{these arguments}. There are many extensions and generalisations of the framework \change{as elaborated in} \cite{Brewka2017AbstractDF}.

\section{Problem Statement}

Among the generalisations of Dung's original framework are  Bench-Capon's value-based argumentation framework (VAF), which was introduced in \cite{capon02}. One of the intentions of the generalisation is to allow individual arguments to be associated with underlying values that contain a preferential ordering relation. This ordering then allows the arguments of the more highly-preferred values to withstand attacks from arguments with a lower value.

Another extension of Dung's original framework is Corsi and Ferm\"uller's semi-abstract argumentation framework (SAF), which was presented in \cite{corfer17lori}. This framework assigns propositional formulae to individual arguments (thus the arguments are not completely abstract anymore, hence the name). This enables defining attack principles (that make implicit attacks explicit) and a consequence relation that only requires arguments and attack relations (referring neither to the truth values nor to the interpretations in the usual sense).

This thesis explores whether the two frameworks can be united into a formalism that would incorporate the intuitions and main ideas of both of the aforementioned frameworks. \change{A} complex moral dilemma concerning the ethics of autonomous vehicles is then examined using \change{Dung's and Bench-Capon's frameworks.} 

\section{Aim of the Work}

The aim of the work is to create a new argument framework, one that combines the concepts from Bench-Capon's value-based argumentation framework, and Corsi and Ferm\"uller's semi-abstract argumentation framework. This is done by assigning values to individual logical formulae.

\change{Dung's original argumentation framework and Bench-Capons value-based framework} are then used for the analysis of a moral dilemma in a manner that highlights the differences between \change{the two formalisms. Furthermore, the obstacles which prevent representing the dilemma using Corsi and Ferm\"uller's semi-abstract argumentation framework and the newely-introduced semi-abstract value-based framework are discussed.}

\section{Structure of the Work}

Section two presents the relevant concepts from Dung's argumentation framework, Bench-Capon's value-based argumentation framework, and Corsi and Ferm\"uller's semi-abstract argumentation framework. Along with definitions, original examples are provided. In section three the new framework is introduced along with \change{an example encompassing the newly-introduced concepts}. The moral dilemma is presented in section four. Firstly, the dilemma is analysed in the terms of Dung's argumentation framework and the VAF. The dilemma is then expanded as to showcase the advantages that the VAF offers over the original framework. Lastly, \change{it is shown that propositional logic is not expressive enough for a satisfying representation of the moral dilemma, which prevents the creation of formulae which would provide the basis for the application of SAF and the new framework}.

\chapter{State of the Art}

In this chapter, the argumentation framework by Dung, value-based argumentation framework by Bench-Capon, and semi-abstract argumentation framework by Fermüller and Corsi \change{are} introduced. The first represents the original proposal for representing argumentation via graphs (where nodes are individual arguments and directed edges are attack relationships) whereas the later two expend the original framework by assigning the individual arguments (i.e. the nodes) values and propositional formulae, respectively.

\section{Argumentation Framework}
\label{afsection}

In 1995 an article by Phan Minh Dung \cite{dung95} introduced a theory of argumentation which was based on accessibility of arguments. On the grounds that the proposed formalisation was very influential in the following decades\footnote{for example: \cite{baroni2011afra}, \cite{dunne2002coherence}, and \cite{egly2010answer}}, it is beneficial to understand the original concepts introduced in this article.

\subsection{Argumentation Examples}

Let us analyse the following dialogue:

\begin{example}
\label{conversationExample} \footnote{This example and the analysis are inspired by an example from \cite{dung95}} \\
\emph{Person A:} You are to be blamed for the fact that we have not sat down for a coffee yet, since you did not call me. \\
\emph{Person B:} No, the blame is entirely on you, because you did not called me. \\
\emph{Person A:} Yes, but I believe I deserve that you call me first.
\end{example}

In this discussion we can recognise three arguments. The first argument of person A (let us call that argument \setword{a1}{a1}) assigns the blame for not meeting earlier to person B. Person B's response (let us call it argument \setword{b}{b}) attacks the argument \ref{a1}. Dung \cite{dung95} recognises that, if this dialogue were to end at this point, neither side could reasonably claim the victory because both arguments use the same reasoning and therefore cannot ``beat'' the opposing argument without simultaneously weakening their own position. 

However the second argument by person A (let us call it \setword{a2}{a2}) uses a different excuse in passing the blame (namely that person A deserved to be called first). If the conversation were to end at this point, person A could claim the victory because argument \ref{a2} defeats argument \ref{b}, making argument \ref{a1} acceptable.
\\ \\
Before explaining the framework, let us provide another, more complex, example. Each sentence (argument) will be named in the parentheses at the end of the line.

\begin{example}
\label{opposingValuesExample} During a walk, two men \change{see} a beggar asking for money. \\
\emph{Emmanuel:} I feel it is my moral duty to give someone in need what they ask for. We should give money to this person, since that is what he asks for. \emph{(\change{A}rgument name: \setword{e1}{e1})} \\
\emph{Jeremy:} By giving this man money, you may actually cause him more harm. We should give him food instead. \emph{(\setword{j1}{j1})} \\
\emph{Emmanuel:} But if this man needs medicine or clothing, food won't help him. \emph{(\setword{e2}{e2})} \\
\emph{Jeremy:} If we give him food, at least we cannot cause him harm. \emph{(\setword{j2}{j2})}
\end{example}

In this discussion it is not immediately clear who would be the winner if the conversation were to be finished at this point. Although Emmanuel was the one to finish the conversation, it is not clear if giving the man food instead of money really is the better choice. The reason for this will be discussed more closely in section \ref{vafsection}.

In the subsection \ref{afconcepts} we will formally analyse examples \ref{conversationExample} and \ref{opposingValuesExample}.

\subsection{Argumentation Framework and Extensions}
\label{afconcepts}

Definitions from this subsection follow the ones originally proposed in \cite{dung95}. The first concept that ought to be formally defined is \emph{argumentation framework}. 

\begin{definition}
\label{afDef}
An \emph{argumentation framework} is a pair \\
\centerline{\(AF = \langle AR, attacks\rangle\)}
where \(AR\)  is a set of arguments and \(attacks\) is a binary relation on \(AR\), i.e.\\ \(attacks\subseteq AR \times AR\).
\end{definition}

At this point it is important to clarify that in \cite{dung95} an argument is \change{represented as} an abstract entity \change{with no explicit} internal structure. Most extensions and other versions of argumentation frameworks (including value-based argumentation framework in section \ref{vafsection} and semi-abstract argumentation, section \ref{safsection}) are building upon this \change{aspect} of Dung's definition. Furthermore, it is also important to say that all attacks are equally strong. This is another point that is often defined differently in other versions of argumentation framework.
\\
Argumentation framework can also be represented using directed graphs\footnote{for example:  \cite{rahwan2009argumentation}, \cite{corfer18fuzzyArgumentation}, and \cite{capon02}}, where each node represents an argument and an edge from \(n1\) to \(n2\) represents the relation \(attacks(n1, n2)\).
\\ \\
Applying definition \ref{afDef} to arguments from example \ref{conversationExample} results in the framework below: \\
\centerline{\(AR = \langle\{\ref{a1}, \ref{b}, \ref{a2}\}, \{(\ref{a1},\ref{b}), (\ref{b},\ref{a1}), (\ref{a2},\ref{b})\}\rangle\).}

This framework has the following graph:

    \begin{center}    
        \fbox{
        \begin{minipage}{0.9\textwidth}
            \begin{center}
            \begin{tikzpicture}[rn/.style={circle, draw=black!60, fill=lightgray!5, very thick, minimum size=8mm},]
            \label{graph1}
            
            \node[rn]    (g1a1)      {\ref{a1}};
            \node[rn]    (g1b)   [right=of g1a1] {\ref{b}};
            \node[rn]    (g1a2)    [below=of g1b] {\ref{a2}};
             
            \draw[->,thick] (g1a1.north) to[bend left] (g1b.north);
            \draw[->, thick] (g1b.south) to[bend left] (g1a1.south);
            \draw[->, thick] (g1a2.north) to[bend right] (g1b.east);
            \end{tikzpicture}
            \end{center}
        \end{minipage}
        }
    \end{center}

On the other hand, the argumentation framework and argumentation graph for the discussion from example \ref{opposingValuesExample} would be:
\\
\centerline{\(AR = \langle\{\ref{e1}, \ref{j1}, \ref{e2}, \ref{j2}\}, \{(\ref{e1},\ref{j1}), (\ref{j1},\ref{e1}), (\ref{e2},\ref{j1}), (\ref{j2},\ref{e2})\}\rangle\).}
\\
      \begin{center}
        \fbox{
        \begin{minipage}{0.9\textwidth}
            \begin{center}
            \begin{tikzpicture}[rn/.style={circle, draw=black!60, fill=lightgray!5, very thick, minimum size=8mm},]
            \label{graph2}
            
            \node[rn]    (g2e1)      {\ref{e1}};
            \node[rn]    (g2j1)   [right=of g2e1] {\ref{j1}};
            \node[rn]    (g2e2)    [below=of g2j1] {\ref{e2}};
            \node[rn]    (g2j2)    [right=of g2e2] {\ref{j2}};
             
            \draw[->,thick] (g2e1.north) to[bend left] (g2j1.north);
            \draw[->, thick] (g2j1.south) to[bend left] (g2e1.south);
            \draw[->, thick] (g2e2.north) to[bend right] (g2j1.east);
            \draw[->, thick] (g2j2.west) to (g2e2.east);
            \end{tikzpicture}
            \end{center}
       \end{minipage}
       }
    \end{center}
    
Furthermore, it should be defined what it means for arguments to be \emph{conflict-free}.

\begin{definition}
\label{conflictFreeDef}
A set S of arguments is said to be \emph{conflict-free} if there are no arguments A and B in S where A attacks B.
\end{definition}

In the previous definition as well as in the rest of this section, an argument \(A\) \emph{attacks} an argument \(B\) (i.e. \(B\) \emph{is attacked by} \(A\)) means that \(attacks(A,B)\) holds.  Furthermore an arbitrary set of arguments \(S\) attack an argument \(A\) (i.e. \(A\) \emph{is attacked by} \(S\)) if there is at least one argument in \(S\) that attacks \(A\).

Dung \cite{dung95} further defines concepts of \emph{acceptable} and \emph{admissible} sets of arguments.

\begin{definition}
\label{acceptableDef}
An argument \(A \in AR\) is \emph{acceptable} with respect to a set of arguments \(S\)  if for each argument \(B \in AR\) it holds that if \(B\) attacks \(A\) then \(B\) is attacked by \(S\).
\end{definition}

\begin{definition}
\label{admissibleDef}
A conflict-free set of arguments \(S\) is \emph{admissible} if each argument in \(S\) is acceptable with respect to \(S\).
\end{definition}

After defining an admissible set, we can also define a \emph{preferred extension}.

\begin{definition}
\label{prefferedExtensionDef}
A \emph{preferred extension} of an argumentation framework \(AF\) is a maximal (with respect to set inclusion) admissible set of \(AF\).
\end{definition}

Let us apply definition \ref{prefferedExtensionDef} to our example \ref{conversationExample}. Conflict-free sets in the example are \(\{\}\), \(\{a1\}\), \(\{a2\}\), \(\{b\}\), \(\{a1, a2\}\). Admissible sets are \(\{\}\), \(\{a2\}\) and \(\{a1, a2\}\). It is important to note that \(\{b\}\) is not admissible, because \(attacks(a2,b)\) holds, yet there is no argument in set \(\{b\}\) which attacks \(a2\). Since a preferred extension is a maximal admissible set, it is clear that \(\{a1,a2\}\) is the only preferred extension of our example. Lastly, Dung \cite{dung95} defines the \emph{stable extension}.

\begin{definition}
\label{stableExtensionDef}
A conflict-free set of arguments \(S\) is called a \emph{stable extension} if \(S\) attacks each argument which does not belong to \(S\).
\end{definition}

It is clear that \(\{a1,a2\}\) is the only stable extension from example \ref{conversationExample}. In example \ref{opposingValuesExample}, we have two stable extensions: \(\{j1,j2\}\) as well as \(\{e1, j2\}\). 

\subsection{Recognised Shortcomings of the Framework}
\label{afrecognised}

In \change{\cite{dung95}, Dung} recognises the need for further research and modifications of his proposed argumentation framework. Specifically, he recognises the need for:

\begin{enumerate}
    \item Showing that some arguments have different strengths;
    \item Solving the problem of self-defeating arguments (one example of an AF with a self-defeating argument A is: \(\langle\{A,B\},\{(A,A),(A,B)\}\rangle\));
    \item Further study about the architecture and development of argumentation systems.
\end{enumerate}

Section \ref{vafsection} presents one approach to solving the first problem in this list.

\section{Value-Base\change{d} Argumentation Framework}
\label{vafsection}

In 2002 Trevor Bench-Capon \cite{capon02} presents his value-based argumentation framework. The main change that he introduces to Dung's theory is changing \change{argument representation} from purely abstract entities to entities that have values. These values enable arguments to acquire a relative strength in relation to other arguments. As values of arguments depend on a certain value system, Bench-Capon's framework enables us to differentiate between objectively true and subjectively true conclusions. Objectively true conclusions are arguments that are acceptable for a given set \(S\) of arguments regardless of the preferred value system for \(S\), whereas subjectively true conclusions are arguments that are acceptable for a given set \(S\) of arguments for some, but not necessarily for all value systems (further discussion on this issue is in subsection \ref{vafproperties}).

\change{S}ubsection \ref{vafconcepts} provides an insight into formal definitions of value-based argumentation framework, value systems\change{,} and subjectively and objectively true conclusions and offers a simple example for clarifying these definitions.

\subsection{Value-Based Argumentation Framework Concepts}
\label{vafconcepts}

Bench-Capon \cite{capon02} defines his \emph{value-based argumentation framework} in the following manner:

\begin{definition}
\label{vafdef}
A \emph{value-based argumentation framework} (VAF) is a 5-tuple:

$$ VAF = \langle AR, attacks, V, val, valpref \rangle $$

Where \(AR\), and \(attacks\) are as in standard argumentation framework (definition \ref{afDef}). \(V\) is a non-empty set of  values, \(val\) is a function which maps from  elements of \(AR\) to elements of \(V\), and \(valpref\) is a preference relation on \(V \times V\). Function \(valpref\) has to have the following properties: \emph{transitivity}, \emph{irreflexivity}, and \emph{asymmetry}.
\end{definition}

As we have seen in example \ref{opposingValuesExample}, arguments are based on different values and thus have different strengths. Therefore it is necessary to define when an attack is successful.

\begin{definition}
\label{vafdefeatsdef}
\emph{An argument \(A \in \  AF\) defeats an argument \(B \in \ AF\)} if the following two conditions stand: \(attacks(A,B)\) and not \(valpref(val(B), val(A))\).
\end{definition}

In the definition above, \(valpref(val(B), val(A))\) denotes that the value of the argument B is preferred over \change{the} value of A. Another way of clarifying the ordering of values (which is also used in chapters \ref{myArgChapter} and \ref{chapterApplication}) is by using the inequality signs ($>$ and $<$) as in the numbers theory (in this particular case, the notation would be $val(B) > val(A)$ or $val(A) < val(B)$). \change{Now} we can define when an argument is acceptable. Informally, an argument A is acceptable if all arguments from S that could defeat A are defeated by some argument in S.

\begin{definition}
\label{vafacceptabledef}
An argument \(A \in AR\) is \emph{acceptable} with respect to set of arguments \(S\), marked as \(acceptable(A,S)\) if: \\
\( \forall x \  (( x \in \ S \ \land \ defeats(x,A)) \implies \exists y \ (( y \in  \  S) \ \land \  defeats(y,x)))\). 
\end{definition}

The definition of conflict-free set of arguments for a VAF is less strict than the definition we saw in AF (definition \ref{conflictFreeDef}). Here, a set of arguments can still be conflict-free even if it contains attacks, provided that no argument is defeated.

\begin{definition}
\label{vafconflictfreedef}
A set \(S\) of arguments is \emph{conflict-free} if: \\
\( \forall x \  \forall y \ (( x \in \ S \ \land \ y \in \ S) \implies (\neg attacks(x,y) \lor valpref(val(y),val(x))))\).
\end{definition}

A set of arguments is admissible if it is conflict-free and each argument is acceptable with respect to the whole set.

\begin{definition}
\label{vafadmisibledef}
A conflict-free set of arguments \(S\) is \emph{admissible} if: \\
\( \forall x (x \in S \implies acceptable(x,S))\)
\end{definition}

Lastly, the definition of the preferred extension ought to be redefined with respect to multiple values.

\begin{definition}
\label{vafpreferreddef}
A set of arguments \(S\) in an argumentation framework is a \emph{preferred extension} if it is a maximal (with respect to set inclusion) admissible set of \(AR\).
\end{definition}

Definition of a stable extension is the same as in Dung's original argumentation framework (definition \ref{stableExtensionDef}).

\begin{definition}
\label{vafstabledef}
A conflict-free set of arguments \(S\) is a \emph{stable extension} if \(S\) attacks each argument in \(AR\) which does not belong to S.
\end{definition}

\subsection{Reconsidering Argumentation Example}
\label{vafreconsidering}

We can now reconsider the example \ref{opposingValuesExample}. The reason why it was not clear if Jeremy truly could be considered the winner is because arguments often rely on different underlying values. 

Arguments j1 and j2 represent \change{simplified} views of utilitarian ethics, where \change{(for the purposes of this demonstration)} actions are encouraged if they will lead to the greatest amount of happiness\footnote{\change{The utilitarian ethics are more complex than this ad hoc interpretation.} More information on utilitarianism available at \url{https://www.utilitarianism.com/bentham.htm} (last visited on: 2018-02-27)}. On the other hand, arguments e1 and e2 follow a \change{value system somewhat reminiscent of} Kantian ethics, which \change{(again, for the purposes of this demonstration)} states that a person should act according to some moral principles even if such action will not necessarily lead to the greatest amount of happiness \footnote{\change{This is an ad hoc simplification of the undelying ethical principles.} More information on Kantian ethics available at \url{http://www.csus.edu/indiv/g/gaskilld/ethics/kantian\%20ethics.htm} (last visited on: 2018-02-27)}. Therefore, neither friend is able to convince the other because each prefers his own moral system (i.e. value) more. 

One way to graphically represent different values (which was also proposed \change{in} \cite{capon02}) is to use different colours for nodes representing different values. Here, we can use white for Kantian ethics and grey for ethics of utilitarianism. Alternatively (or, as in our case, additionally), an additional symbol for the value of the argument can be written under its name. In  our case we will use K and U.

    \begin{center}
    \fbox{
    \begin{minipage}{0.9\textwidth}
        \begin{center}
            \begin{tikzpicture}[rn/.style={circle, draw=black!60, fill=lightgray!5, very thick, minimum size=8mm},]
            \label{graph2vaf}
            
            \node[rn, align=center]    (g2e1)      {\ref{e1} \\ K};
            \node[rn, align=center, fill = lightgray]    (g2j1)   [right=of g2e1] {\ref{j1} \\ U};q
            \node[rn, align=center]    (g2e2)    [below=of g2j1] {\ref{e2} \\ K};
            \node[rn, align=center, fill = lightgray]    (g2j2)    [right=of g2e2] {\ref{j2} \\ U};
             
            \draw[->,thick] (g2e1.north) to[bend left] (g2j1.north);
            \draw[->, thick] (g2j1.south) to[bend left] (g2e1.south);
            \draw[->, thick] (g2e2.north) to[bend right] (g2j1.east);
            \draw[->, thick] (g2j2.west) to (g2e2.east);
            \end{tikzpicture}
        \end{center}
    \end{minipage}
    }
    \end{center}

Formally, this VAF would be defined as: 
\begin{center}
    {\(A = \langle\{\ref{e1}, \ref{j1}, \ref{e2}, \ref{j2}\}, \{(\ref{e1},\ref{j1}), (\ref{j1},\ref{e1}), (\ref{e2},\ref{j1}), (\ref{j2},\ref{e2})\}, \{U, K\}, val, valpref\rangle\)}
\end{center}
where \\
\centerline{\(val(arg) = 
    \begin{cases*}
        U \  | \  arg \in \{\ref{j1},\ref{j2}\} \\
        K \  | \  arg \in \{\ref{e1},\ref{e2}\} \\
    \end{cases*}
\)}

and either \(valpref = \{K < U\}\) (if we prefer utilitarianism over Kantian ethics) or \(valpref = \{U < K\}\).

We will now analyse case when \(valpref = \{K < U\}\). In this case: \(defeats(j2,e2)\), \(\neg defeats(e2, j1)\), \(defeats(j1, e1)\) and \( \neg defeats(e1, j1) \). Thus, \(acceptable(j2,A)\) and \(acceptable(j1,A)\). Because of \(defeats(j2,e2)\) and \(defeats(j1, e1)\), A is not conflict-free. The set \(\{j1, j2\}\) is then the the only preferred extension.

However, if \(valpref = \{U < K\}\) then \(val(\ref{j2}) < val(\ref{e2})\), thus \ref{j2} will not defeat e2, although \(attacks(j2,e2) \in A\). Namely, in this case we have \(\neg defeats(j2,e2)\), \(defeats(e2, j1)\), \(\neg defeats(j1, e1)\)\change{,} and \( defeats(e1, j1) \). From there, we get relations  \(acceptable(e2,A)\) and \(acceptable(e1,A)\), but we also get \(acceptable(j2,A)\), since no argument attacks j2. Once again, A is not conflict-free (this time because of \(defeats(e2,j1)\) and \(defeats(e1, j1)\)). Note however, that the subset of this VAF \(S_1 = \langle\{\ref{e2},\ref{j2}\},\) \(\{(\ref{j2},\ref{e2})\},\) \(\{U, K\},\) \(\{val(\ref{j2}) = K,\) \(val(\ref{e2}) = I\},\) \(\{U < K\}\rangle\) would also be conflict-free (definition \ref{vafconflictfreedef}), although it contains an attack. Preferred extension in this case is \(\{e1,e2,j2\}\). This means that a rational agent would always have to agree that buying the beggar food will certainly not cause him harm (unlike giving him money). Nevertheless, a rational agent could still decide that giving him money is the better choice, but only if it prefers Kantian ethics. 

As we have already seen in subsection \ref{afconcepts}, we have two stable extensions: \(\{\ref{j1},\ref{j2}\}\) as well as \(\{\ref{e1}, \ref{j2}\}\), regardless of \(valpref\). This shows us that, just like in the original argumentation framework, not all preferred extensions are stable.

\subsection{Properties of Dichromatic Value-Based Argumentation Frameworks}
\label{vafproperties}

We have seen in subsection \ref{vafreconsidering} that using VAF can provide a better understanding of a disagreement if we provide multiple values that reflect multiple viewpoints. In this subsection we will only look at VAFs that contain multiple values (formally, \(|V|>1\)). These VAFs are called \emph{polychromatic}, because if each value is represented with a colour (as in subsection \ref{vafreconsidering}), the graph will contain multiple colours. We will additionally emphasise the properties of \emph{dichromatic} frameworks, which contain exactly two values (\(|V|=2\)).

One important property of the argumentation framework proposed by Bench-Capon \cite{capon02} is the following:

\begin{property}
Given an order of values, a polychromatic cycle in VAF has a unique, non-empty preferred extension.
\end{property}

This is very important, as in AF, odd-cycles (which \change{Bench-Capon in \cite{capon02} dubs} paradoxes) will have the empty set as their preferred extension and even cycles (which \change{are called dilemmas in \cite{capon02}}) will have two preferred extensions, \change{as was shown in} \cite{capon02, dunne2002coherence}. However, if there are no monochromatic cycles in a VAF, each VAF will have a \emph{non-empty and unique} preferred extension.

\begin{comment}
However, if there are no monochromatic cycles in a VAF, each VAF will have a \emph{non-empty and unique} preferred extension and Capon offers the following algorithm for calculating that extension: 

EXTEND(AR, attacks):
\begin{enumerate}
    \item \(S := \{s \langle AR: \forall y (\neg defeats(y,s))\}\)
    \item \(R := \{ r \langle AR: \exists s S for which defeats(s,r(\}\)
    \item \(If S = \emptyset then return S and finish halt\)
    \item \(AR' := AR / (S \leftarrow R)\)
    \item This makes no sence. Page 4 is the rest
\end{enumerate}
\end{comment}

Another advantage of VAF is the ability to differentiate between \emph{objectively acceptable} and \emph{subjectively acceptable} arguments. An argument is objectively acceptable if it can be acceptable regardless of value order. Furthermore, if there is at least one value order under which an argument is acceptable the argument is subjectively acceptable, otherwise it is \emph{indefensible} \cite{capon02, dunne2002coherence}. As we have seen in subsection \ref{vafreconsidering}, argument \ref{j2} is objectively acceptable, and all other arguments are subjectively acceptable (as each is true in exactly one of the two value orders).

Bench-Capon\cite{capon02} presents a way in which it can be determined if an argument is objectively acceptable, subjectively acceptable or indefensible without the need to analyse the whole framework. In order to explain that approach, we need to define the concept of a \emph{chain}.

\begin{definition}
\label{vafchaindef}
An \emph{argument chain} C in a VAF is a set of arguments \(\{a_1, \dots , a_n\}\) such that:
\begin{enumerate}
    \item \(\forall a_i \forall a_j (( a_i \in C \land a_j \in C) \implies val(a_i) = val(a_j))\);
    \item \(a_1\) has no attacker in \(C\);
    \item For all \(a_i \in C\): if \(i > 1\), then \(a_i\) is attacked and the sole attacker of \(a_i\) is \(a_{i-1}\).
\end{enumerate}
\end{definition}

Although definition \ref{vafchaindef} was introduced in this form by Bench-Capon in his paper \cite{capon02}, in the same paper the author implicitly redefines the \change{third} rule so that \change{the} arguments that are attacked by more than one other argument can still be seen as the last \change{element} of a longer chain (rather than considering all such arguments as chains with only one argument). In order to formalise Bench-Capon's implicit changes of definition \ref{vafchaindef} that are necessary for the property \ref{vafpreferredicy}, we will provide a new definition that will be used for the remainder of this thesis \change{(including property \ref{dichromProperties} and table \ref{vafpathstable})}.

\begin{definition}
\label{vafchainNEWdef}
An \emph{argument chain} in a VAF, C is a set of arguments \(\{a_1, \dots , a_n\}\) such that:
\begin{enumerate}
    \item \(\forall a_i \forall a_j (( a_i \in C \land a_j \in C) \implies val(a_i) = val(a_j))\);
   % \item \(\forall b\  (b \in VAF\  \land\  val(b) = val(a_1)\  \land\  attacks(b,a_1) \if \exists C' \  (C'\) is a chain in VAF \(\ \land\  b \in C'\  \land\  b\) is the last argument of \(C'\) because of the rule 4\()\);
    \item \(a_1\) has no attacker in \(C\);
    \item For all \(a_i \in C\): if \(i > 1\), then \(a_i\) is attacked by \(a_{i-1}\) and there is no other argument in \(C\) which attacks \(a_i\);
    \item If \(a_i \in C\), where \(i > 1\) is attacked by an argument that is not in \(C\), then \(a_i\) is the last argument in chain \(C\);
    \item If for \(a_i \in C\) the rule 4. applies and if \(a_i\) attacks an argument \(b\) with \(val(a_i) = val(b)\), then \(b\) will be the first argument of another chain.
\end{enumerate}
\end{definition}

By adding rule 5, we have ensured that if an argument \(a\) is the last element of one chain because of multiple attacks on \(a\), \(a\) will not be considered the first element of the next chain. The first element instead is the argument which would have been \(a\)'s successor had \(a\) not been attacked by \change{another} argument. Defining parity is also very important. An argument is said to be \emph{odd} (or \emph{even}) if it is an odd (or even) element in a chain. The first element of a chain has the position \change{one} and it is, therefore, odd. We can see that definition \ref{vafchainNEWdef} allows each argument to be a member of multiple chains. In this case the parity will be defined as described in property \ref{vafmultiplechainsprop}.

\begin{property}
\label{vafmultiplechainsprop}
An argument should be seen as an element of multiple chains if one of the following two conditions are fulfilled:
\begin{enumerate}
    \item Argument \(a\) is the last argument in multiple chains (because it is attacked by multiple arguments of the same value). If \(a\) attacks another element, that attack will be seen as an attack of an even (i.e. odd) chain if \(a\) is even in at least one chain (i.e. \(a\) is odd in all chains).
    \item Argument \(a\) attacks multiple arguments with the same value.% If \(a\) is an even element in at least one of the nodes, its parity is even in all chains. Otherwise, \(a\)'s parity in all chains is odd. 
\end{enumerate}
\end{property}

Now we can quote another property from \cite{capon02} that determines how a preferred extension can be found in a dichromatic cycle.
\\
\begin{property}
\label{vafpreferredicy}
The preferred extension of a dichromatic cycle comprises:
\begin{enumerate}
    \item the odd numbered arguments of all chains preceded by an even chain;
    \item the odd numbered arguments of chains with the preferred value;
    \item the even numbered arguments of all other chains.
\end{enumerate}
\end{property}

In property \ref{vafpreferredicy}, rule 1 includes objectively acceptable arguments, while rules 2 and 3 comprise subjectively acceptable arguments. Bench-Capon \cite{capon02, dunne2002coherence} also includes a property that describes when an argument can be objectively or subjectively acceptable or indefensible in any dichromatic VAF (not necessarily a cycle). 
\\
\begin{property}
\label{dichromProperties}
In a dichromatic VAF:
\begin{enumerate}
    \item An argument is \emph{indefensible} if it is an even numbered member of any chain preceded only by even chains; or if it is an even numbered member of a chain attacked by an odd chain and is directly attacked by an odd chain;
    \item An argument is \emph{objectively acceptable} if it is only an odd numbered argument of a chain preceded only by even chains;
    \item An argument is subjectively acceptable otherwise.
\end{enumerate}
\end{property}

These rules can also be summarised in the form of five ``paths'' presented in table \ref{vafpathstable}.

%\begin{table}[ht]
%\centering
%\caption{Paths for a dichromatic VAF}
%\label{vafpathstable}
%\begin{tabular}{|l|l|l|}
\begin{longtable}{|l|l|l|}
\hline
\rowcolor[HTML]{32CB00}
Path 1 & Objectively Acceptable  & \(odd\  \land\  \neg CAOC\) \\ \hline
\rowcolor[HTML]{FCFF2F} 
Path 2 & Subjectively Acceptable & \(odd\  \land\  CAOC \) \\ \hline
\rowcolor[HTML]{FCFF2F} 
Path 3 & Subjectively Acceptable & \(even\  \land\  CAOC\  \land\  \neg AAOC\) \\ \hline
\rowcolor[HTML]{FE0000} 
Path 4 & Indefensible            & \(even\  \land\  CAOC\  \land\  AAOC \) \\ \hline
\rowcolor[HTML]{FE0000} 
Path 5 & Indefensible            & \(even\  \land\  \neg CAOC \) \\ \hline
\caption{Paths for a dichromatic VAF}
\label{vafpathstable}
\end{longtable}
%\end{tabular}
%\end{table}

The meaning of abbreviations in table \ref{vafpathstable}, is as follows:
\begin{itemize}
    \item odd: Argument is on an odd position in its chain
    \item even: Argument is on an even position in its chain
    \item CAOC: The Chain (of the argument) is Attacked by an Odd Chain
    \item AAOC: The Argument is (directly) Attacked by an Odd Chain
\end{itemize}

\change{The conclusions from property \ref{dichromProperties}, which are summarised in table \ref{vafpathstable}, are provided in this form in \cite{capon02}. In their justification, Bench-Capon provides a number of examples from which it is unambiguous that the used notion of a chain follows definition \ref{vafchainNEWdef}, and not definition \ref{vafchaindef} (which is the only definition of a chain to be found in the paper).}

\section{Semi-Abstract Argumentation Framework}
\label{safsection}

We saw in section \ref{vafsection} that Bench-Capon \change{changed the representation of} arguments from completely abstract entities to entities that have values. A more widespread approach to removing the abstractness from Dung's framework is to use logical formulae to describe the \change{support and claim} of \change{an argument}.

For example, in \cite{Besnard2009} Philippe Besnard and Anthony Hunter present \emph{argumentation based on classical logic}. They represent arguments as pairs $\langle\Phi, \alpha\rangle$ where $\Phi$ is a minimal consistent set of formulae that proves $\alpha$. In other words, an argument consists of a support ($\Phi$) and a claim ($\alpha$) which can be inferred from its support by some inference method. A  counterargument  for  an  argument $\langle\Phi, \alpha\rangle$ is defined as an  argument $\langle\Psi, \beta\rangle$ where the claim $\beta$ contradicts the support $\Phi$. In \cite{Besnard2009} the claim and the support are defined as formulae in classical propositional logic. There are many variants of this approach that use other logics for defining the claim and the support, as elaborated in \cite{master18}. 

Recently Fermüller and Corsi defined \emph{semi-abstract argumentation} (or SAF) in \cite{corfer17lori}. The SAF can be seen, to an extent, as a framework whose basis represents a midpoint between Dung's abstract argumentation and Besnard's arguments based on classical logic. Thus defined arguments have a claim that is represented via a propositional formula \change{whereas the support is abstract. The fact that the claim is explicitly given while support is not is reflected in the name of the framework.} This approach enabled the authors to define a logical consequence operator based on logical attack principles as well as to tackle the issue of making implicit attacks explicit. In the remained of this chapter we will further examine SAFs.

\subsection{Semi-Abstract Argument\change{ation} and Attack Principles}
\label{safconcepts}

The main intuition behind semi-structured argumentation is to see the framework as Dung's ordinary abstract argumentation framework (section \ref{afconcepts}) with a propositional formula attached to each argument. In this sense, we need to define the set of propositional formulae $\mathcal{PL}$. \change{All} definitions in this subsection as well as in subsection \ref{safConsequence} (with the exception of definition \ref{safDef}) follow the ones from \cite{corfer17lori}.

\begin{definition}
\label{defsafPropositionalFormulae}
Let $\mathcal{PV}$ be an infinite set of propositional variables. Then we define the set of propositional formulae $\mathcal{PL}$ over $\mathcal{PV}$ by

$$\mathcal{F\ ::=\ F \lor F\ |\ F \land F\ |\ F \supset F\ |\ \lnot F\ |\ PV}$$

where $\mathcal{F}$ is a meta variable that stands for a propositional formula.\end{definition}

Semi-structured argumentation is formally defined in \cite{corfer17lori} as follows.
\begin{definition}
\label{safDefGraph}
A \emph{semi-abstract argumentation frame} (SAF) is a directed graph $(AR, R_\rightarrow)$ where each node $A \in AR$ is labelled by a formula of $\mathcal{PL}$ representing the claim of argument $A$ and the edges $R_\rightarrow$ represent the attack relation between the arguments.
\end{definition}

One can see that this definition references the usual way Dung's abstract argumentation is defined in recent papers (further examples include \cite{egly2010answer} and \cite{bistarelli2009solving}) -- where the framework is defined in terms of graphs. In this representation, nodes are arguments and directed edges are attack relations. A definition that does not reference graphs (and is thus more inline with definitions \ref{afDef} and \ref{vafdef}) is also possible and it relies on defining claims via a labelling function, as in \cite{master18}.

\begin{definition}
\label{safDef}
A \emph{semi-abstract argumentation framework} is a triple:

$$ SAF = \langle AR, attacks, \lambda  \rangle$$

Where $AR$ and $attacks$ are as in standard argumentation framework (definition \ref{afDef}) and $\lambda$ is a labelling function $\lambda: \ AR \rightarrow \mathcal{PL}$ which labels each argument $a \in AR$ with a propositional formula $\mathcal{F} \in \mathcal{PL}$.
\end{definition}

We can read $A\rightarrow B$ as ``the argument (with claim) A attacks the argument (with claim) B''. Furthermore, if an argument with claim C does not attack an argument with claim D, we will mark this as $C \not\rightarrow D$.

Defining arguments in this way has the advantage of allowing us to see implicit attacks in a framework. For example, if one argument attacks an argument with the claim $a$, then it is reasonable to assume that the same argument implicitly attacks all other arguments with ``stronger'' claims $a \land \mathcal{F}$, with $\mathcal{F} \in \mathcal{PL}$. In order to make the implicit attacks explicit as in the example above, logical attack principles have to be defined. In \cite{corfer17lori} the fundamental principle states that if an argument $A$ attacks another argument $B$, then $A$ also attacks all other arguments $C$ from which $B$ can be inferred (here and in the remainder of this chapter it will be implicitly assumed that $A, B, C,\dots \in AR$).

\begin{definition}
\label{safGenAttackPrinciple}
\emph{General attack principle:} 

\change{(A) If $F \rightarrow G$ and $G' \models G$ then $F \rightarrow G'$.}
\end{definition}

Intuitively the general attack principle can be understood as the principle which states that if an agent disagrees with a conclusion, it also has to disagree with all premises that lead directly to that conclusion. One should also note that the implication used in \ref{safGenAttackPrinciple} does not have to be the implication used in predicate logic. Indeed, this is one of the points where the framework introduced by Corsi and Fermüller \cite{corfer17lori} offers great flexibility. It is only important that the argument $G'$ follows directly from $G$ in some sense. As explained in \cite{corfer17lori}, if one adopts the definition of ``$\models$'' in minimal logic from \cite{troelstra2000basic}, then the general attack principle entails transparent instances of \change{(A)}.

\begin{definition}
\label{attackprinciplesA}
    \emph{Transparent instances of \change{(A)}:}

\textbf{(A.$\land$)} If $F \rightarrow A$ and $F \rightarrow B$, then $F \rightarrow A \land B.$
    
\textbf{(A.$\lor$)} If $F \rightarrow A \lor B$ then $F \rightarrow A$ and $F \rightarrow B$.

\textbf{(A.$\supset$)} \vthree{If $F \rightarrow A \supset B$} then $F \rightarrow B$
\end{definition}

The proof that these transparent instances follow directly from \change{(A)} in minimal logic is shown in proposition 1 of \cite{corfer17lori}. The principle (A.$\supset$) may appear less intuitive th\vthree{a}n the remaining two. This was recognised by Corsi and Fermüller and they have provided an alternative in the form of the principle (B.$\supset$) which corresponds to ``basic inference principle about implicit premises in logical consequence claims'' \cite{corfer17lori}. The authors further note that (B.$\supset$) holds in a wide range of logics and reflects the implication in relevance logic from \cite{dyrkolbotn2013formal}. 

\begin{definition}
\label{attackprinciplesBimpl}
\emph{Relevant entailment principle (B.$\supset$)}

\textbf{(B.$\supset$)} If $F \not\rightarrow A$ and $F \rightarrow B$ then $F \rightarrow A \supset B$.
\end{definition}

The attack principle for negation is given on the premise that if an \change{F} attacks another argument \change{A}, then \change{F} does not also attack the negation of \change{A}. Although this reasoning provides a clear intuitive meaning of the principle, the authors of \cite{corfer17lori} also noted that it cannot be inferred from \change{(A)} in minimal logic from \cite{CM_1937__4__119_0} because this logic treats $\bot$ as any other constant symbol.

\begin{definition}
\label{attackprinciplesBneg}
\emph{Negation attack principle}

\textbf{(B.$\lnot$)} If $F \rightarrow A$ then $F \not\rightarrow \lnot A$.
\end{definition}

Lastly, the authors of \cite{corfer17lori} define the inverse attack principles for negation attack principle, relevant entailment principle and the first two transparent instances of \change{(A)}. Note that the inverse property (A.$\supset$) \change{is} not defined.

\begin{definition}
\label{attackprinciplesC}
\emph{Inverse attack properties}

\textbf{(C.$\land$)} If $F \rightarrow A \land B$ then $F \rightarrow A$ or $F \rightarrow B$.

\textbf{(C.$\lor$)} If $F \rightarrow A$ and $F \rightarrow B$ then $F \rightarrow A \lor B$.

\textbf{(C.$\supset$)} If $F \rightarrow A \supset B$ then $F \not\rightarrow A$ or $F \rightarrow B$.

\textbf{(C.$\lnot$)} If $F \not\rightarrow A$ then $F \rightarrow \lnot A$.
\end{definition}

In order to be able to define argumentative consequence in SAF (subsection \ref{safConsequence}), we want to explicitly define the set of attack principles $\mathcal{AP}$. 

\change{
\begin{definition}
\label{safActualAP}
Let $\mathcal{AP}$ be any (proper) subset of attack principals (A.$\land$), (A.$\lor$), (A.$\supset$), (B.$\supset$), (B.$\lnot$), (C.$\land$), (C.$\lor$), (C.$\supset$), (C.$\lnot$), as defined in \ref{attackprinciplesA}, \ref{attackprinciplesBimpl}, \ref{attackprinciplesBneg}, and \ref{attackprinciplesC}.
\end{definition}
}

\change{One $\mathcal{AP}$ which was investigated in \cite{corfer17lori} is CAP. It includes} all attack principles defined above with the exception of \change{(A)} and (A.$\supset$).

\begin{definition}
\label{safCAP}
Let \change{CAP} be the set of attack principles containing (A.$\land$), (A.$\lor$), (B.$\supset$), (B.$\lnot$), (C.$\land$), (C.$\lor$), (C.$\supset$), (C.$\lnot$), as defined in \ref{attackprinciplesA}, \ref{attackprinciplesBimpl}, \ref{attackprinciplesBneg}, and \ref{attackprinciplesC}.
\end{definition}

\change{N}ote that some of the principles in \change{CAP} entail that any SAF S containing them is infinite (e.g. (A.$\land$)). As there are numerous cases where having a finite number of arguments is necessary, one can \emph{relativise} each of the attack principles defined above so that they only apply to a set of formulae $\Gamma$. As one can see in an example from \cite{master18}, the relativised attack principle for (A.$\land$) would state: For every $A, B, C \in \Gamma:\ A \rightarrow B$ or $A \rightarrow C$ implies $A \rightarrow B \land C$, if $B \land C \in \Gamma$. Like in the paper \cite{corfer17lori}, we too assume that the attack principles are relativised to the appropriate set of formulas. 

\subsection{Argumentative Consequence}
\label{safConsequence}
Corsi and Fermüller in \cite{corfer17lori} also introduced a \change{definition of} consequence relation dubbed argumentative consequence. A significant aspect of this \change{definition} is that it requires neither truth values nor interpretations in the usual (Tarskian) sense. The \change{leading} premise is that the attacking argument's claim can be seen as a countermodel for the attacked arguments claim.

\change{As the authors of \cite{corfer17lori} note, in} order for \change{this definition of consequence} to be logically acceptable, the underlying SAFs need to be rich enough to contain subformulae of the occurring formulae. For example, a SAF $S$ that contains arguments $a$ and $a \lor b$ would not be rich enough because it contains no argument with the claim $b$. Formally, argumentative consequence requires \emph{logical closure of arguments} as defined below.

\begin{definition}
\label{safClosure}
A $SAF$ $S$ is \emph{logically closed} with respect to the set of formulae $\Gamma$ if all subformulae of formulae in $\Gamma$ occur as claims of some argument in $S$.
\end{definition}

\change{Logical closure is the necessary prerequisite for making implicit arguments explicit in a logically acceptable manner. Thus the closure of the underlying language is established and the argumentative consequence can be defined.}

\begin{definition}
\label{safConsequenceDefinition}
$F$ is an \emph{argumentative consequence} of (the claims of) arguments $A_1, \dotsc, A_n \in AR$, where $AR$ are the arguments with respect to an $SAF$ $S$ (c.f. definitions \ref{safDef} or \ref{safDefGraph}), denoted as $A_1, \dotsc , A_n \models^S_{AR} F$ if all arguments in $S$ that attack $F$ also attack one of arguments $A_1, \dotsc A_n$.
\end{definition}

\change{It is significant to recognise the finding of the authors of \cite{corfer17lori} that if we were to see arguments as counter-models and $S$ were to contain all relevant counter-models (c.f. definition \ref{safClosure}), then ordinary logical consequence and argumentative consequence as defined \ref{safConsequenceDefinition} would coincide because the counter-models always invalidate some premise of the conclusion.} As an illustration, let argument $F$ be the claim that fast food burgers are healthy, argument $A_1$ be the claim that fast food burgers contain vegetables, and argument $A_2$ be that meals containing vegetables are usually healthy. In this example it holds that $A_1, A_2 \models F$ because every argument that attacks $F$ clearly has to attack $A_1$ or $A_2$. \change{As in \cite{corfer17lori}, definition \ref{safConsequenceGeneralisationDefinition} further generalises the} definition \ref{safConsequenceDefinition} to sets of SAFs.

\begin{definition}
\label{safConsequenceGeneralisationDefinition}
For a set of SAFs $\mathcal{S}$ we say that $A_1, \dotsc , A_n \models^\mathcal{S}_{AR} F$ if $A_1, \dotsc , A_n \models^S_{AR} F$ for every $S \in \mathcal{S}$.
\end{definition}

\change{In the investigation of the relationship between argumentation and logics in \cite{corfer17lori}, the definition of the argumentative $\mathcal{AP}$-consequence below was of special importance. Some of the results of this investigation are provided in subsection \ref{safAPinvestigationSection}.}

\begin{definition}
\label{safAPconsequence}
Let $\mathcal{AP}$ be a set of attack principles (\change{c.f. definition \ref{safActualAP}}). $F$ is called an \emph{argumentative $\mathcal{AP}$-consequence} of $A_1,\dotsc , A_n$, denoted as $A_1, \dotsc , A_n \models_{AR}^{\mathcal{AP}} F$, if $A_1, \dotsc , A_n \models_{AR}^{S} F$ for every $SAF$ $S$ that is logically closed with respect to $\{A_1, \dotsc , A_n,\ F\}$ and moreover satisfies all (appropriately relativised) attack principles in $\mathcal{AP}$. 
\end{definition}

Lastly, let us define argumentative consequence for sets of SAFs and finite sets of formulae, as in \cite{corfer17lori}. 

\begin{definition}
\label{safConsequenceDefintionsSets}
Let $\Gamma$ and $\Delta$ be finite sets of formulae and let $S$ be an $SAF$. We say that $\Delta$ is the \emph{argumentative consequence of $\Gamma$ under $S$}, written $\Gamma \models^{S}_{AR} \Delta$, if all arguments in $S$ that attack every $F \in \Delta$ attack at least some $G \in \Gamma$.
\end{definition}

\begin{definition}
\label{safConsequenceDefintionsSetsAndSetsOfSAFs}
Let $\Gamma$ and $\Delta$ be finite sets of formulae and let $\mathcal{S}$ be a set of SAFs. We say that $\Gamma \models^{\mathcal{S}}_{AR} \Delta$ if $\Gamma \models^{S}_{AR} \Delta$ for every $S \in \mathcal{S}$.
\end{definition}

\subsection{\change{Justification of Attack Principles}}
\label{safAPinvestigationSection}

In \cite{corfer17lori} the authors have proven the soundness and completeness of Gentzen's \cite{gentzen1935untersuchungen} classical sequent calculus for the case when the attack principles used for the argumentative $\mathcal{AP}$-consequence are the ones from \change{CAP (definition \ref{safCAP})}. The \change{corresponding proofs} can be seen in chapter 5 of \cite{corfer17lori}. \change{Therefore, if all of the attack principles in CAP were to be accepted, the logically closed SAFs would correspond to the classical logic over the language $\mathcal{PL}$ (definition \ref{defsafPropositionalFormulae}). This would essentially reduce reduce formal argumentation to propositional logic.}

\change{However, there are reasons against accepting all arguments from CAP. For example, the principle $(A.\land)$ states that if an argument attack a conjecture of arguments it must also attack at least one of the conjectured arguments. But let us now consider the case $F \rightarrow A \land \neg A$. Clearly, we would expect a rational agent to oppose that some claim and its direct negation are simultaneously true because that would be a logical contradiction. But that does not unavoidably mean that the agent would necessarily need to attack either of the claims. A concrete example would be the claim ``octopuses have fewer limbs than squids and octopuses do not have fewer limbs than squids''. Whereas no knowledge about the marine animals is required for refuting the contradictory conjunction, one does need to know the number of limbs of the two cephalopods in order to attack either alternative on its own.}

\change{Continuing with this line of thought, we can see can understand the intuition behind refuting $(C.\neg)$. Namely, it is often not reasonable to anticipate that every argument that does not attack another argument $A$ has to attack its negation $\neg A$. Specifically, just because there is no reason for attacking the argument ``octopuses have fewer limbs than squids'' does mean that the alternative ``octopuses do not have fewer limbs than squids'' has to be attacked. Corsi and Ferm\"uller \cite{corfer17lori} also note that $(C.\neg)$ should be contrasted with the more reasonable principle $(B.\neg)$ which just states that no argument should attack $A$ and $\neg A$ simultaneously.}

\change{Lastly, let us consider the attack principle $(B.\supset)$. It states that if $F \rightarrow B$ and $F \not\rightarrow A$ then $F \rightarrow A \supset B$. An intuitive example where this principle could be seen as problematic is the case where argument B has the claim ``the real-estate prices are at the historic maximum'', and the argument A could have the claim that the ``housing market is in an economic bubble''. A rational agent could potentially attack B (by e.g. providing a point in the past where the prices is even higher) and also not attack the presumption that the housing market is an economic bubble (due to insufficient data to do so). However, this agent could still believe that $A \supset B$ because it could be possible that a bubble would indeed lead to the highest prices in the history.}

\change{Admittedly, the arguments above are far from formal justification. Furthermore, the argument against $(B.\supset)$ could be seen as less convincing than the ones agains $(A.\land)$ and $(C.\neg)$. The authors of \cite{corfer17lori} did not rely on pre-theoretic reasoning (like the ones above). Instead, the authors found ``a simple formal interpretation of attack involving logically compound claims, that supports some, but not all of the attack principles in CAP'' \cite{corfer17lori}. To this end \cite{corfer17lori} applies standard modal logic and a class of Kripke Interpretations (c.f. \cite{grossi2010argumentation}) with the weaker condition of seriality. The notation for the corresponding class Kripke interpretations is $\mathcal{D}$. The complete description of this formalism and the proofs of (non-)justifiability is available in chapter 6 of \cite{corfer17lori}.}

\change{In this manner, Corsi and Ferm\"uller's \cite{corfer17lori} investigation concludes that the attack principles $(A.\land)$, $(A.\lor)$, $(B.\neg)$, $(C.\land)$, and $(C.\supset)$ are $\mathcal{D}$-justified. On the other hand, $(B.\supset)$, $(C.\lor)$, and $(C.\neg)$ are not $\mathcal{D}$-justified.}

\change{This formal justification is then used for the the following definition of $\mathcal{AP}$.}

\change{
\begin{definition}
\label{safMAP}
Let \change{MAP} be the set of attack principles containing $(A.\land)$, $(A.\lor)$, $(B.\neg)$, $(C.\land)$, and $(C.\supset)$, as defined in \ref{attackprinciplesA}, \ref{attackprinciplesBneg}, and \ref{attackprinciplesC}.
\end{definition}
}

\change{Unlike CAP (c.f. defintion \ref{safCAP}, MAP suggests that the ``logic of argumentation'' (as dubbed in \cite{corfer17lori}) is indeed weaker than classical logic. The authors of \cite{corfer17lori} therefore proclaim that MAP is more plausible than CAP. As stated in the beginning of this subsection, $\models_{AR}^{CAP}$ (c.f. definition \ref{safAPconsequence}) induces classical logical consequence. In contrast, in chapter 7 of \cite{corfer17lori} it is shown that $\models_{AR}^{MAP}$  matches the sequent calculus LM (as defined in e.g. \cite{negri2005proof}).}

\chapter{Suggested \change{New Framework}}
\label{myArgChapter}

\change{This chapter introduces} a hybrid argumentation framework that is based on Corsi and Fermüller's semi-abstract argumentation framework (section \ref{safconcepts}) and Bench-Capon's value-based argumentation framework (section \ref{vafsection}). The argumentation framework associates individual formulae with values and adapts the definitions of argumentative consequences accordingly.

\section{Semi-Abstract Value-Based Argumentation Framework}
\label{myArgSection}

Before defining the proposed framework, we need to define an ordering with the smallest value which will be used in this context.

\begin{definition}
\label{myAFprefWithZeroDef}
Let $V$ be a non-empty set whose elements are called values. Then we call $valpref$ a \emph{proper preference relation over $V$} if it is a relation over $V \times V$ such that it is \emph{transitive}, \emph{irreflexive}, \vthree{and} \emph{asymmetrical}.
\end{definition}

As was the case with previous pure value-based argument framework, we can define semi-abstract value-based argumentative framework both with and without referencing graphs. It is important to note that a propositional formula is defined in the same manner as it was the case for the SAF (definition \ref{defsafPropositionalFormulae}).

\begin{definition}
\label{myAFdef}
A \emph{semi-abstract value-based argumentation framework} (SVAF) is a 5-tuple:

$$ VAF = \langle AR, attacks, \lambda, V, val, valpref \rangle $$

Where $AR$ and \(attacks\) are as in standard argumentation framework (definition \ref{afDef}), while $\lambda$ is a labelling function $\lambda: \ AR \rightarrow \mathcal{PL}$ which labels each argument $a \in AR$ with a propositional formula $\mathcal{F} \in \mathcal{PL}$. Furthermore, \(V\) is a non-empty set of values, \(val\) is a function which maps from elements of \(AR\) to elements of \(V\), and \(valpref\) is a proper preference relation over $V$ (c.f. definition \ref{myAFprefWithZeroDef}).
\end{definition}

The same framework \change{could also be defined via a directed graph $(AR,R_\rightarrow$) where each vertex $A \in AR$ is labelled by a formula of $\mathcal{PL}$ representing the claim of argument $A$, as well as a value from a non-empty set of values $V$. Then, the edges $R_\rightarrow$ represent the attack relations between the arguments while $val$, $V$, and $valpref$ are all as in the definition above.}

\change{Every argument in} $AR$ is associated with \change{a} formula \change{from} $PL$ and \change{a} value \change{from} $V$. \change{For instance, say }that $f \in PL$ is the claim \change{and} $v \in V$ is the value of \change{an} argument \change{-- then we can mark this argument as a pair $(f, v)$ and write $(f, v) \in AR$. However, one should not forget that -- just as in the case of SAF (c.f. section \ref{safsection}) -- the claims of arguments are to be seen as occurrences of formulae, rather than the formulae per se. In other words, it is possible to have multiple arguments with the same claim and even the same claim-value pair. In order to avoid such ambiguities (or to avoid writing a formula every time an argument is referenced, which can be problematic in case of more complex formulae), we can also assign labels to individuals arguments. Thus, for instance, we could use a label such as $A$ (or any other potentially meaningful name) in order to represent the aforementioned argument $(f,v)$.}  If we have another attack $B \in AR$ \change{(where $B$ is another label)}, which is attacked (\change{alternatively,} not attacked) by the argument $A$, we write this as $A \rightarrow B$ (i.e. $A \not\rightarrow B$ if the attack does not exist). 

\change{Similarly as in the case of SAF, SVAF can be used for defining consequence relations. Furthermore, logical closure remains a reasonable requisite for making implicit arguments explicit. As we can see in definition \ref{svafClosure}, \vthree{closure is defined on the level of individual values. For example, a framework containing arguments with claim-value pairs $\{(a\land b, v), (a, v), (b, w)\}$ (with $a, b \in \mathcal{PV}$ and $v, w \in val$) is not closed because of the absence of argument $(b,v)$.}}

\vthree{
\begin{definition}
\label{svafClosure}
Let $S$ be a SVAF with values $val = \{v_1,\dotsc,v_n\},\ n \in \mathbb{N}$. We say that $S$ is is logically closed with respect to the value $v_i,\ i \in \{1,\dotsc,n\}$ and a set of formulae $\Gamma_i$ if all subformulae of formulae in $\Gamma_i$ occur as claims of some arguments with value $v_i$ in $S$. Furthermore, we say that $S$ is \emph{logically closed} with respect to the sets of formulae $\Gamma_1,\dotsc\Gamma_n$ if $S$ is logically closed with respect to all values $v_1,\dotsc,v_n$ and sets  $\Gamma_1,\dotsc\Gamma_n$.
\end{definition}}

\change{Now we can define attack principles and sets of attack principles in a manner analogues to what was originally introduces by Corsi and Fermüller. \vthree{Similarly as in the case of logical close, values of the arguments have to be considered. In  definitions \ref{MYGenAttackPrinciple} and \ref{MYattackprinciples}, it is assumed that $f, g, g', a, b \in \mathcal{PL}$ and $v, w \in val$ of some given SVAF.}}

\begin{definition}
\label{MYGenAttackPrinciple}
\vthree{\emph{General attack principle:}}

\change{(A) If $(f,v), \rightarrow (g,w)$ and $g' \models g$ then $(f,v) \rightarrow (g',w)$.}
\end{definition}

\begin{definition}
\label{MYattackprinciples} \vthree{\emph{Attack principles}:}

\textbf{(A.$\land$)} If $(f,v) \rightarrow (a,w)$ and $(f,v) \rightarrow (b,w)$, then $(f,v) \rightarrow (a \land b,w).$
    
\textbf{(A.$\lor$)} If $(f,v) \rightarrow (a \lor b,w)$ then $(f,v) \rightarrow (a,w)$ and $(f,v) \rightarrow (b,w)$.

\textbf{(A.$\supset$)} If $(f,v) \rightarrow (a \supset b,w)$ then $(f,v) \rightarrow (b,w)$

\textbf{(B.$\supset$)} If $(f,v) \not\rightarrow (a,w)$ and $(f,v) \rightarrow (b,w)$ then $(f,v) \rightarrow (a \supset b,w)$.

\textbf{(B.$\lnot$)} If $(f,v) \rightarrow (a,w)$ then $(f,v) \not\rightarrow (\lnot a,w)$.

\textbf{(C.$\land$)} If $(f,v) \rightarrow (a \land b,w)$ then $(f,v) \rightarrow (a,w)$ or $(f,v) \rightarrow (b,w)$.

\textbf{(C.$\lor$)} If $(f,v) \rightarrow (a,w)$ and $(f,v) \rightarrow (b,w)$ then $(f,v) \rightarrow (a \lor b,w)$.

\textbf{(C.$\supset$)} If $(f,v) \rightarrow (a \supset b,w)$ then $(f,v) \not\rightarrow (a,w)$ or $(f,v) \rightarrow (b,w)$.

\textbf{(C.$\lnot$)} If $(f,v) \not\rightarrow (a,w)$ then $(f,v) \rightarrow \lnot (a,w)$.
\end{definition}

The intuition as well as the logical background for all of the attack principles above is therefore inline with that of Corsi and Fermüller's SAF. \vthree{The definitions of sets of attack principles also follow the reasoning from \cite{corfer17lori}.} 

\begin{definition}
\label{MYActualAP}
\vthree{Let $\mathcal{AP}$ be any (proper) subset of attack principals (A.$\land$), (A.$\lor$), (A.$\supset$), (B.$\supset$), (B.$\lnot$), (C.$\land$), (C.$\lor$), (C.$\supset$), (C.$\lnot$), as in definition \ref{MYattackprinciples}.}
\end{definition}
\begin{definition}
\label{MYCAP}
\vthree{Let \change{CAP} be the set of attack principles containing (A.$\land$), (A.$\lor$), (B.$\supset$), (B.$\lnot$), (C.$\land$), (C.$\lor$), (C.$\supset$), (C.$\lnot$), as in definition \ref{MYattackprinciples}.}
\end{definition}
\begin{definition}
\label{MYMAP}
\vthree{Let \change{MAP} be the set of attack principles containing $(A.\land)$, $(A.\lor)$, $(B.\neg)$, $(C.\land)$, and $(C.\supset)$, as in definition \ref{MYattackprinciples}.}
\end{definition}

\vthree{Lastly,} we will also assume that only the relativised formulae (i.e. just the subformulae
of the original formulae in $\Gamma$) are to be considered in order to be able to work with finite
sets of arguments. We will now clarify the introduced concepts and definitions with the example below.

\begin{example}
 \label{mySvafExamplePart1}
Let us assume that we have a SVAF with two values (\change{we will name the two values} $prog$ and $trad$) whose arguments contain the following claim-value pairs: \vthree{ $\{(\neg a, prog)$, $(a \lor b, prog)$, $(b \supset c, prog)$, \vthree{$(a, trad)$,} $(b \land d, trad)$, $(b,trad)$, $(d,trad)\}$}, where $a, b, c,$ and $d$ are propositional variables. Furthermore, let the following attacks be given: \vthree{$\{(\neg a, prog) \rightarrow (a, trad)$, $(a, trad) \rightarrow (\neg a, prog)$, $(\neg a, prog) \rightarrow (a \lor b, prog)$, $(a \lor b, prog) \rightarrow (\neg a, prog)$, $(a \lor b, prog) \rightarrow (d, trad)$, $(b \land d, trad) \rightarrow (b \supset c, prog)$, $(b \supset c, prog) \rightarrow (b,trad)$, $(b \supset c, prog) \rightarrow (d,trad)\}$. This SVAF is represented graphically below}. 

    \begin{center}
    \fbox{
    \begin{minipage}{0.9\textwidth}
        \begin{center}
            \begin{tikzpicture}[rn/.style={circle, draw=black!60, fill=lightgray!5, very thick, minimum size=8mm},]
            \label{svafGraph1}
            
            \node[rn, align=center]    (notaProg)      {$\lnot a$ \\ prog};
            \node[rn, align=center]    (aorbProg)   [right=of notaProg] {$a \lor b$ \\ prog};
            \node[rn, align=center] (bimpliescProg) [right=of aorbProg] {$b\supset c$ \\ prog};
            
            \node[rn, align=center, fill = lightgray]    (aTrad)    [below=of notaProg] {$a$ \\ trad};
            \node[rn, align=center, fill = lightgray]    (banddTrad)    [below=of bimpliescProg] {$b\land d$ \\ trad};
             
            \node[rn, align=center, fill = lightgray]    (bTrad)    [right=of banddTrad] {$b$ \\ trad};
            \node[rn, align=center, fill = lightgray]    (dTrad)    [right=of bimpliescProg] {$d$ \\ trad};

            \draw[->, thick] (notaProg) to[bend right] (aorbProg);
            \draw[->, thick] (aorbProg) to[bend right] (notaProg);
            \draw[->, thick] (notaProg) to[bend right] (aTrad);
            \draw[->, thick] (aTrad) to[bend right] (notaProg);
            \draw[->, thick] (aorbProg.north) to[bend left] (dTrad);
            
            \draw[->,thick] (banddTrad) to (bimpliescProg);
            \draw[->, thick] (bimpliescProg) to (bTrad);
            \draw[->, thick] (bimpliescProg) to (dTrad);

            \end{tikzpicture}
        \end{center}
    \end{minipage}
    }
    \end{center}

\vthree{This argumentation framework is not closed, which prevents applying attack principles. So, let us insert the arguments needed for the logical closure: $(a,prog)$, $(b,prog)$, and $(c,prog)$. Note that we had to add $(a,prog)$ even though $(a,trad)$ was already given (c.f. definition \ref{svafClosure}).}

Now we can make implicit attacks explicit by applying \change{MAP} attack principles (definition \vthree{\ref{MYMAP}}). \change{We will use MAP (and not CAP or some other $AP$) given that the remained of the attack principles can be seen as too strong, as discovered by Corsi and Fermüller and discussed in subsection \ref{safAPinvestigationSection}. For easier readability, we will now reiterate the statements of the principles contained in MAP.}

\begin{align}
    (A.\land)&\textrm{: If }\vthree{(f,v)} \rightarrow \vthree{(a,w)}\textrm{ and }\vthree{(f,v)} \rightarrow \vthree{(b,w)}\textrm{, then }\vthree{(f,v)} \rightarrow \vthree{(a }\land \vthree{b,w)}\textrm{.} \notag\\ 
    (A.\lor)&\textrm{: If } \vthree{(f,v)} \rightarrow \vthree{(a} \lor \vthree{b,w)}\textrm{ then }\vthree{(f,v)} \rightarrow \vthree{(a,w)}\textrm{ and }\vthree{(f,v)} \rightarrow \vthree{(b,w)}\textrm{.} \notag\\ 
    (B.\neg)&\textrm{: If }\vthree{(f,v)} \rightarrow \vthree{(a,w)}\textrm{ then }\vthree{(f,v)} \not\rightarrow \vthree{(}\lnot\vthree{a,w)}\textrm{.} \notag\\ 
    (C.\land)&\textrm{: If }\vthree{(f,v)} \rightarrow \vthree{(a} \land \vthree{b,w)}\textrm{ then }\vthree{(f,v)} \rightarrow \vthree{(a,w)}\textrm{ or }\vthree{(f,v)} \rightarrow \vthree{(b,w)}\textrm{.} \notag\\ 
    (C.\supset)&\textrm{: If }\vthree{(f,v)} \rightarrow \vthree{(a} \supset \vthree{b,w)}\textrm{ then }\vthree{(f,v)} \not\rightarrow \vthree{(a,w)}\textrm{ or }\vthree{(f,v)} \rightarrow \vthree{(b,w)}\textrm{.} \notag
\end{align}

\vthree{By applying $(A.\land)$ onto $(b \supset c, prog) \rightarrow (b,trad)$ and $(b \supset c, prog) \rightarrow (d,trad)\}$, we can infer $(b \supset c, prog) \rightarrow (b \land d,trad)$. From $(\neg a, prog) \rightarrow (a \lor b, prog)$ by principle $(A.\lor)$, we infer $(\neg a, prog) \rightarrow (a, prog)$ and $(\neg a, prog) \rightarrow (b, prog)$. However, from the attack of $(\neg a, prog)$ on $(a \lor b, prog)$ by applying $(A. \lor)$ we \emph{could not have} inferred $(\neg a, prog) \rightarrow (a, trad)$ (had this attack not already been given) because the values of the arguments differ (c.f. definition \ref{MYattackprinciples}). The principle $(B.\neg)$ makes $(a, trad) \not\rightarrow (a,prog)$ explicit from $(a, trad) \rightarrow (\neg a,prog)$. Had the attacks $(b \supset c, prog) \rightarrow (b,trad)$, $(b \supset c, prog) \rightarrow (d,trad)\}$ not already been given, we could have inferred that at least one of them has to exist by applying $(C.\land)$ to the newly-inferred $(b \supset c, prog) \rightarrow (b \land d,trad)$ -- this showcases why the principle $(C.\land)$ is seen as the inverse to $(A.\land)$. Lastly, we have principle $(C.\land)$ which can be applied on the attack $(b \land d, trad) \rightarrow (b \supset c, prog)$ to infer $(b \land d, trad) \not\rightarrow (b, prog)$ or $(b \land d, trad) \rightarrow (c, prog)$. In summary, the following relations have been inferred: $\{ (b \supset c, prog) \rightarrow (b \land d,trad)$, $(\neg a, prog) \rightarrow (a, prog)$, $(\neg a, prog) \rightarrow (b, prog)$, $(a,trad) \not\rightarrow (a, prog)\}$. Furthermore, it has to be the case that $(b \land d, trad) \not\rightarrow (b, prog)$ or $(b \land d, trad) \rightarrow (c, prog)$.} 

\vthree{The closed argumentation framework with the inferred attacks can be seen below. The fact that $(b \land d, trad) \rightarrow (c, prog)$ is not certain is illustrated by using a dashed edge.}

    \begin{center}
    \fbox{
    \begin{minipage}{0.9\textwidth}
        \begin{center}
            \begin{tikzpicture}[rn/.style={circle, draw=black!60, fill=lightgray!5, very thick, minimum size=8mm},]
            \label{svafGraph2}
            
            \node[rn, align=center]    (notaProg)      {$\lnot a$ \\ prog};
            \node[rn, align=center]    (aorbProg)   [right=of notaProg] {$a \lor b$ \\ prog};
            
            \node[rn, align=center] (aProg) [below=of aorbProg] {$a$ \\ prog};
            \node[rn, align=center] (bProg) [right=of aorbProg] {$b$ \\ prog};
            \node[rn, align=center] (cProg) [below=of bProg] {$c$ \\ prog};
            
            \node[rn, align=center] (bimpliescProg) [right=of bProg] {$b\supset c$ \\ prog};
            
            \node[rn, align=center, fill = lightgray]    (aTrad)    [below=of notaProg] {$a$ \\ trad};
            \node[rn, align=center, fill = lightgray]    (banddTrad)    [below=of bimpliescProg] {$b\land d$ \\ trad};
             
            \node[rn, align=center, fill = lightgray]    (bTrad)    [right=of banddTrad] {$b$ \\ trad};
            \node[rn, align=center, fill = lightgray]    (dTrad)    [right=of bimpliescProg] {$d$ \\ trad};

            \draw[->, thick] (notaProg) to[bend right] (aorbProg);
            \draw[->, thick] (aorbProg) to[bend right] (notaProg);
            \draw[->, thick] (notaProg) to[bend right] (aTrad);
            \draw[->, thick] (aTrad) to[bend right] (notaProg);
            \draw[->, thick] (aorbProg.north) to[bend left] (dTrad);
            
            \draw[->,thick] (banddTrad) to[bend right] (bimpliescProg);
            \draw[->, thick] (bimpliescProg) to (bTrad);
            \draw[->, thick] (bimpliescProg) to (dTrad);
            
            \draw[->,thick] (bimpliescProg) to[bend right] (banddTrad);
            \draw[->,thick] (notaProg) to (aProg);
            \draw[->,thick] (notaProg.north) to[bend left] (bProg);
            
            \draw[->, thick, dashed] (banddTrad) to (cProg);
            
            \end{tikzpicture}
        \end{center}
    \end{minipage}
    }
    \end{center}

\end{example}

\subsection{Objective and Subjective Argumentative Consequence}
\label{myArgumentativeConsequenceSection}

In order to define argumentative consequence in the SVAF, we have to revisit the way in which attacks relate to values. This will be done in a similar manner as was the case in VAF (c.f. definition \ref{vafdefeatsdef}). 

\begin{definition}
\label{myDefeatsDefintion}
An argument A \emph{defeats} another argument B from a SVAF if A attacks B and it is not the case that the value of B is preferred over the value of A.
\end{definition}

Now, we can define the argumentative consequence relation which will be based on the defeat relation.

\begin{definition}
\label{myConsequenceDefinition}
$F$ is a \emph{subjective argumentative consequence} of (the claims of) arguments $A_1, \dotsc, A_n \in AR$, where $AR$ are the arguments with respect to a SVAF $S$ (definition \ref{myAFdef}), denoted as $A_1, \dotsc , A_n \models^S_{SUB} F$ if all arguments in $S$ that defeat $F$ also defeat one of arguments $A_1, \dotsc A_n$ for at least one preference \vthree{ordering}. 

If $F$ is a subjective argumentative consequence of (the claims of) arguments $A_1, \dotsc, A_n \in AR$ for all possible value preference \vthree{orderings}, then it is called \emph{objective argumentative consequence} and denoted as $A_1, \dotsc , A_n \models^S_{OBJ} F$.
\end{definition}

\begin{definition}
\label{myAPconsequence}
Let $\mathcal{AP}$ be a set of attack principles. $F$ is called a \emph{subjective (i.e. objective) argumentative $\mathcal{AP}$-consequence} of $A_1,\dotsc , A_n$, denoted as $A_1, \dotsc , A_n \models_{SUB}^{\mathcal{AP}} F$ (i.e. $A_1, \dotsc , A_n \models_{OBJ}^{\mathcal{AP}} F$), if $A_1, \dotsc , A_n \models_{SUB}^{S} F$ (i.e. $A_1, \dotsc , A_n \models_{OBJ}^{S} F$) for every $SAF$ $S$ that is logically closed with respect to $\{A_1, \dotsc , A_n,\ F\}$ and moreover satisfies all (appropriately relativised) attack principles in $\mathcal{AP}$. 
\end{definition}

\begin{definition}
\label{myConsequenceDefintionsSets}
Let $\Gamma$ and $\Delta$ be finite sets of formulae and let $S$ an SVAF. We say that $\Delta$ is the \emph{subjective (i.e. objective) argumentative consequence of $\Gamma$ under $S$}, written $\Gamma \models^{S}_{SUB} \Delta$ ($\Gamma \models^{S}_{OBJ} \Delta$), if all arguments in $S$ that defeat every $F \in \Delta$ defeat at least some $G \in \Gamma$ for at least one value preference \vthree{ordering (i.e. all value preference orderings)}.
\end{definition}

\begin{definition}
\label{myConsequenceDefintionsSetsAndSetsOfSAFs}
Let $\Gamma$ and $\Delta$ be finite sets of formulae and, and let $\mathcal{S}$ be a set of SVAFs. We say that $\Gamma \models^{\mathcal{S}}_{SUB} \Delta$ (i.e. $\Gamma \models^{\mathcal{S}}_{OBJ} \Delta$) if $\Gamma \models^{S}_{SUB} \Delta$ (i.e. $\Gamma \models^{S}_{\vthree{OBJ}} \Delta$) for every $S \in \mathcal{S}$.
\end{definition}

The intended level for the argumentative $AP$-consequence (definition \ref{myAPconsequence}) is not a specific argumentation framework, but rather all possible frameworks that satisfy $AP$. In that sense, definition \ref{myConsequenceDefinition} can be seen as a technical preliminary. Nevertheless, it might be useful to provide subjective and objective argumentative consequences of individual arguments with respect to SVAF from example \ref{mySvafExamplePart1} in order to provide a concrete illustration of the introduced concepts. 

\begin{example}
\label{mySvafExamplePart2}

As the SVAF is logically closed (c.f. definition \change{\ref{svafClosure}}) and the arguments were made explicit using the attack principles from definition \change{\ref{MYattackprinciples}}, the prerequisites for finding objective and subjective argumentative $AP$ consequences are fulfilled \vthree{(where the $AP$ will again correspond to MAP). The SVAF with all arguments and attacks can be seen below (c.f. graphical representation at the end of example \ref{mySvafExamplePart1}). We will observe the case where $(b, \land d, trad) \not\rightarrow (b, prog)$ is included (although not explicitly listed below, together with $(b \land d, trad) \not\rightarrow (b, prog)$), while $(b \land d, trad) \rightarrow (c, prog)$ is omitted.}

\begin{align}
V = \{&prog, trad\} \notag\\
AR = \{&(a, prog),\ (b, prog),\ (c, prog), \notag\\ 
& (\neg a, prog),\ (a \lor b, prog),\ (b \supset c, prog),  \notag\\
& (a, trad),\ (b \land d, trad),\ (b,trad),\ (d,trad)\} \notag\\
attacks = \{&(\neg a, prog) \rightarrow (a, trad), \notag\\
    & (a, trad) \rightarrow (\neg a, prog), \notag\\
    & (\neg a, prog) \rightarrow (a \lor b, prog), \notag\\
    & (a \lor b, prog) \rightarrow (\neg a, prog), \notag\\
    & (a \lor b, prog) \rightarrow (d, trad), \notag\\
    & (b \land d, trad) \rightarrow (b \supset c, prog), \notag\\
    & (b \supset c, prog) \rightarrow (b,trad), \notag\\
    & (b \supset c, prog) \rightarrow (d,trad), \notag\\
    & (b \land d, trad) \rightarrow (c, prog)\} \notag
\end{align}

We begin with arguments that are not attacked (and thus not defeated) by any other argument. \vthree{The only such argument is  $(c, prog)$, since $(b \land d, trad) \rightarrow (c, prog)$ is omitted}. Therefore, \vthree{this} argument \vthree{is an} objective argumentative $AP$-consequences of \vthree{an} empty set.

\vthree{Now the analysis of the non-trivial cases can begin. First, say that $trad$ is preferred over $prog$, i.e. $trad > prog$. In this case, all attacks of arguments with value $prog$ on arguments with value $trad$ will not lead to a defeat and can be disregarded in this part of the analysis. All of the arguments with value $trad$ are then undefeated and are thus subjective argumentative consequences of an empty set. The argument $(\neg a, prog)$ is defeated by $(a, trad)$ and $(a \lor b, prog)$ and it is a subjective consequence of the union of all arguments that these two arguments defeat. Since $(\neg a, prog)$ is indeed only such argument, it can only be a subjective consequence of itself ($(a \lor b, prog)$ does not defeat $(d,trad)$). An analogue case is the argument $(b \supset c, prog)$ because of the attack from $(b \land d, trad)$. Lastly, $(a \lor b, prog)$, $(a,prog)$, and $(b,prog)$ are all subjective consequences of each other because of the attacks from $(\neg a, prog)$.}

\vthree{The case when $prog > trad$ is similar. For example, $(b, trad)$ and $(b \land d, trad)$ are subjective consequences of each other (and of $(d, trad)$), because of the attacks from $(b\supset c, prog)$. Furthermore, the argument $(\neg a, prog)$ is defeated by $(a \lor b, prog)$ which also defeats $(d, trad)$. Therefore, $(\neg a, prog)$ is a subjective argumentative consequence of $(d, trad)$. }

The remaining cases can be seen in table \ref{tabExampleConsequences} below. \vthree{Note that objective argumentative consequences corresponds to the union of the subjective augmentative consequences of a given argument.}

\begin{longtable}{|p{.16\textwidth}|p{.24\textwidth}|p{.24\textwidth}|p{.24\textwidth}|}
\hline
\textbf{Argument} & \textbf{$trad > prog$} & \textbf{$prog > trad$} & \textbf{Objective a.c.} \\ \hline
$(a, prog) $ & $\{(a \lor b, prog),$ $(b, prog)\}$ &  $\{(a \lor b, prog),$ $(b, prog),$ $(a, trad)\}$ & $\{(a \lor b, prog),$ $(b, prog),$ $(a, trad)\}$ \\ \hline
$(b, prog) $ & $\{(a \lor b, prog),$ $(a, prog)\}$ &  $\{(a \lor b, prog),$ $(a, prog),$ $(a, trad)\}$ & $\{(a \lor b, prog),$ $(a, prog),$ $(a, trad)\}$ \\ \hline
$(c, prog) $ & $\{\}$ &  $\{\}$ & $\{\}$ \\ \hline
$(\neg a, prog) $ & $\{(\neg a, prog)\}$ &  $\{(d, trad)\}$ & $\{(\neg a, prog), (d, trad)\}$ \\ \hline
$(a \lor b, prog) $ & $\{(a, prog),$ $(b, prog)\}$ &  $\{(a, prog),$ $(b, prog),$ $(a, trad)\}$ & $\{(a, prog),$ $(b, prog),$ $(a, trad)\}$ \\ \hline
$(b \supset c, prog) $ & $\{(b \supset c, prog)\}$ &  $\{\}$ & $\{(b \supset c, prog)\}$ \\ \hline
$(a, trad) $ & $\{\}$ &  $\{(a \lor b, prog),$ $(a,trad),$ $(b, prog)\}$ & $\{(a \lor b, prog),$ $(a, trad),$ $(b, prog)\}$ \\ \hline
$(b \land d, trad) $ & $\{\}$ &  $\{(b, trad),$ $(d, trad)\}$ & $\{(b, trad),$ $(d, trad)\}$ \\ \hline
$(b,trad) $ & $\{\}$ &  $\{(b \land d, trad),$ $(d, trad)\}$ & $\{(b \land d,$ $trad),$ $(d, trad)\}$ \\ \hline
$(d,trad) $ & $\{\}$ &  $\{(\neg a, prog),$ $(b \land d, trad),$ $(b, trad)\}$ & $\{(\neg a, prog),$ $(b \land d, trad),$ $(b, trad)\}$ \\ \hline
\caption{\vthree{Overview of subjective and objective argumentative consequences}}
\label{tabExampleConsequences}
\end{longtable}

\end{example}

\chapter{Application}
\label{chapterApplication}

In this section we will show how \change{Dung's} argumentation framework \change{and Bench-Capon's} value-based argumentation framework  can be used to derive conclusions in a complex moral dilemma. \vthree{The assumption that a moral dilemma can be satisfyingly represented using (extensions) of Dung's argumentation framework could be disputed. However, \cite{capon02} contains examples that suggest that such a representation indeed is possible. This chapter aims to present another case of this suitability, although the main focus remains on the frameworks themselves rather than on the rapport between the ethics and formal argumentation frames.} 

\vthree{The first section} will \vthree{present} a simpler problem and immediately apply \vthree{abstract} and value-based argumentation framework. \vthree{A}dditional complexity \vthree{will then be integrated in}to the problem in order to showcase the advantages of value-based argumentation framework over Dung's framework. Lastly, \change{an attempt of} \vthree{formalising} the simple problem into \change{propositional} \vthree{logic} \change{will be shown. This attempt will illustrate why the problem cannot be analysed using semi-abstract or semi-abstract value-based argumentation frameworks and propositional logic in a satisfying manner}. 

\section{Moral Dilemma -- Ethics of Autonomous Vehicles}
\label{mdethicsofav}

The example comes from a dilemma that has gained popularity in the recent years thanks to the rapid advancement of autonomous vehicles. \change{In particular, the setting of the problem and the formulation of some of the arguments are primarily inspired by a presentation written by Patrick Lin\footnote{\change{The corresponding animated video (about 4 minutes in length) and the transcript can be seen here: \url{https://www.ted.com/talks/patrick_lin_the_ethical_dilemma_of_self_driving_cars/transcript?language=en\#t-244601} (last visited: 2019-10-27)}}.} Human error is the most common reason for traffic accidents\footnote{More information at: \url{http://cyberlaw.stanford.edu/blog/2013/12/human-error-cause-vehicle-crashes} (last visited: 2017-12-10)}. Autonomous vehicles therefore offer a chance to dramatically lower the total number of accidents by eliminating slow response time and false judgement typical for human drivers. Nevertheless, it is conclusive that some accidents will remain inescapable and in those moments \change{many} ethical questions will \change{become more visible}. 

By the time a human driver becomes aware that an accident is unavoidable it is usually too late for them to consider all possibilities in order to determine the optimal course of action. Instead, the driver’s reaction is usually instinctive and reflexive. Therefore, the driver’s actions are not perceived as conscious choices, \change{making the} raise\change{d} ethical \change{dilemmas less obvious}. 

Autonomous vehicles are radically different in this respect. Since they are operated by a powerful computer, it \change{could} be assumed that autonomous vehicles will be able to “calculate” probable outcomes of \change{multiple} choices that they can make in the moments before \change{the} unavoidable accident. This possibility of making a calculated choice and then acting accordingly is what \change{amplifies} ethical questions, since \change{the} choice of how to evaluate the sacrifices is predetermined \change{long before the moment of crisis}.

The importance of these ethical considerations is especially prominent in the cases where such accident optimisation requires certain “trade-offs” – i.e. a decision on how to evaluate the damage that each possible action in an emergency would have. An extreme example would be a case when all courses of action include a possibility of a serious injury (or even death) for humans. In this rare but possible and important scenario, \change{further} ethical questions arise such as whether the well-being of the driver should be prioritised or, \change{if the} choice has to be made, \change{whether} the car \change{should} hit one child on its left or two elderly citizens on its right. 

\change{It would be wrong to state that the ethical problems that arise in such situations are new. Indeed, the fact that a driver was unable to think in advance does not influence the morality of the choice made. Furthermore, it has to be noted that what is ethically correct does not always correspond to what is legally acceptable. As the aim of this thesis is to showcase the expressivity of the proposed frameworks, the particulars of already-existing legal frameworks will not  be considered in detail.}

Two approaches that are commonly applied to the ethics of autonomous vehicles are the principle of consequentialism and deontological ethics. \change{Unavoidably, the manner in which the values of the two ethics are presented in this thesis will omit numerous integral nuances and details which would be necessary in a rigorous philosophical discourse. This was a conscious decision which was made in order to keep the argument claims and the subsequent formal analysis manageable.}

\change{For the purposes of this analysis, let us say that t}he prime principle of consequentialism is that one should choose the action that leads to the best possible outcome, especially if it is somehow possible to quantify the results\footnote{More information on consequentialism available at: \url{https://www.iep.utm.edu/conseque/} (last visited: 2018-03-02)}. When it comes to autonomous vehicles this would mean that, in emergencies, the vehicle should choose the action that would lead to the least amount of harm (i.e. the smallest number of injured and dead people).

On the other hand, deontological ethics focuses on rightfulness of actions themselves, rather than the rightfulness of consequences of actions\footnote{More information on deontology available at: \url{http://www.philosophybasics.com/branch_deontology.html} (last visited: 2018-02-27)}. \change{In the context of this thesis, we could say that, i}n deontology, the basic principle is that ``right'' actions follow some moral norm, i.e. one should not do anything that is in itself ``wrong'' thing to do. In this sense it could be argued that acting in a way that saves the most number of people could be worse than doing nothing, as there is a distinction between the act of killing and the act of letting die.

\section{Problem Description}
\label{mdproblemdescription}

We will now describe a concrete situation that will serve as a starting point for our argumentation network. Suppose that there is a tunnel containing one lane for each direction. In the right lane, there is a single autonomous car. In the left lane, there is a motorcycle followed by a traditional human-driven car. Unfortunately, the driver of the traditional car is drunk and he suddenly switches to the right lane. Since this situation is happening inside a tunnel and there is not enough time to stop and turn, there are only two courses of action that the autonomous car has:
\begin{enumerate}
    \item Continue straight, in which case both the passenger in the autonomous car and the driver of the traditional car will die.
    \item Turn left and kill the motorcycle rider, but save the life of both the passenger inside the autonomous car and the driver of the traditional car.
\end{enumerate}

The question which this argumentation framework tries to tackle is: ``What is the optimal course of action for the autonomous car?''.

In this framework, we will have two values: consequentialism (C) and deontology (D). Formally: \(V=\{C,D\}\).

\vthree{Before formulating the arguments, it is important to indicate the distinction between two levels of discourse. One level comprises general ethical principles, whereas the other undertakes deciding the optimal course of action in concrete situations. In general, it is not simple to traverse from the first to the the second level. Here, the two values correspond to the level of general ethics, whereas the justifications of individual arguments and attacks correspond to the level of concrete decisions.}

The first argument could be the basic principle of deontological ethics: ``The autonomous car should \change{never take any actions that results in the death of} traffic participants''. We will mark that argument as a. On the other hand, the first argument for the consequentialist side could be: ``The autonomous car has to change lanes if that will result in  the lowest number of casualties'' (this argument will be marked as b). It is clear that in this concrete example, arguments a and b cannot be followed simultaneously and therefore they attack each other. 

Argument a can also be attacked by an argument that follows deontological ethics, as ``If the autonomous car does not change lanes, that can \change{also} be seen as a \change{traffic participant death}, since the passenger in the car would die as a consequence'' (c). However, argument c could be attacked by yet another deontological argument that states that, in this case, ``the passenger would die as a result of inaction\change{, whereas the alternative is much worse because it proposes an action that would directly lead to a participant's death}''(d).

At this point a consequentialist might disagree with d on the basis that, if autonomous vehicles were to \change{regularly} prefer the death of passengers inside them over the other traffic participants, no-one would buy autonomous vehicles. This would mean that the traffic overall would be more dangerous, and this sacrifice would do more harm than good. Therefore, ``an autonomous vehicle should prioritise its own passenger when calculating the optimal damage so that future car buyers do not avoid autonomous vehicles'' (e). \change{Argument e then} attacks argument d, since d says that the death of driver is better than killing someone in the opposite lane\change{. It could be reasonably argued that all deontological arguments (or argument a in particular) would attack this argument. For the sake of keeping the framework manageable, however, we will omit these attacks and focus only on the attack that would come from argument b}, since b states that what matters is saving as many lives in this concrete instance. 

We now add two more consequentialist arguments. The first argument (f) attacks b while being attacked by the second (g). We can define f as ``Moving away and sparing the life of the drunk driver could enable them to cause yet another traffic accident''. And g marks the counterargument to that: ``The non-immanent danger of the drunk driver is not certain, but an additional death in case of moving left is''.

\begin{figure}[h]
    \centering
    \fbox{
    \begin{minipage}{0.9\textwidth}
        \includegraphics[width=13cm]{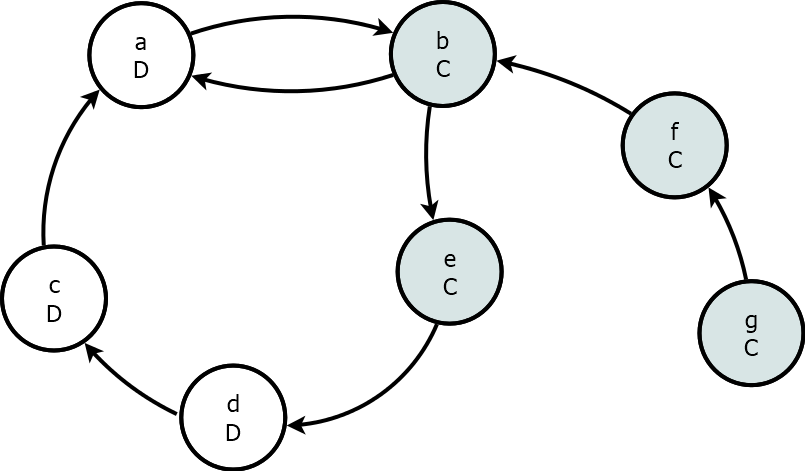}
    \end{minipage}
    }
    \caption{Value-based argumentation framework}
    \label{mdgraph1}
\end{figure}

Let us now represent this argumentation framework graphically in figure \ref{mdgraph1}.

So far, we have two C chains (gfb and be) and one D chain (dca). It is important to note that gfbe is not a single chain because b is attacked by multiple arguments (property \ref{vafmultiplechainsprop}).

Let us now analyse the framework so far using the 5 paths (table \ref{vafpathstable}). We can see that our framework has a cycle (bedca). We start by analysing only arguments that are outside of this cycle - g and f. We can see that g is not attacked, therefore it is objectively acceptable and follows the path 1. Since g objectively acceptable, g attacks f, and g has the same value as f, we can also conclude that f is indefensible. This is to be expected, because f fulfils conditions for path 5. We can therefore conclude that argument f cannot influence b and the cycle can be seen as independent from these two arguments.  

We can thus start with our analysis of the cycle. Let us start with chain containing an odd number (three) of D argument dca. This chain is attacked by a chain with an even number (two) of C arguments. Therefore, argument d will be objectively acceptable since it is in path 1. Argument c is indefensible, since it is on path 5. Lastly, a is only subjectively acceptable, because of \(attacks(b,a)\). Argument a is therefore on path 2.

Lastly, let's look at the chain bc. We have already concluded that f has no influence in b and that a is subjectively acceptable. From there we can conclude that b is also subjectively acceptable, which is to be expected, since it is on path 2. Lastly, e is also subjectively acceptable since it is on path 3.

So to summarise, we have the following situation:
\begin{itemize}
    \item Objectively acceptable: d (D), g (C)
    \item Acceptable if C \(>\) D: b (C)
    \item Acceptable if D \(>\) C: a (D), e (C)
    \item Indefensible: c (D), f (C).
\end{itemize}

The preferred extension in case when deontology is preferred is \{a, d, e, g\}. In case when consequentialism is preferred, the preferred extension is \{b, d, g\}. Regardless of moral preferences, a rational agent would have to accept that the death of the passenger could not be considered a murder (d) as well as that not crashing with the drunk driver does not mean that they will certainly cause another traffic accident (g). On the other hand, the agent would have to dismiss both the argument that not changing lanes also means committing a murder (c) and the argument that it is moral to hit and kill the drunk driver in order to prevent them from causing further harm (f).  

\section{Further Problem Expansion}
\label{mdfurtherproblemanalysis}

Let us now expand our framework somewhat more in order to apply concepts introduced in section \ref{vafsection}. Please note that our priority in this part of the problem description \change{remains} to provide a practical example for concepts in section \ref{vafproperties}, so our understanding of consequentialism and deontology will \change{again omit numerous nuances in order to remain very} flexible.

We start by adding a deontological argument h which will attack c. This argument can be: ``One should follow the law and there is a legal difference between committing a murder and failing to save someone''. The argument h could however be immediately followed by another deontological argument i which would say: ``Laws cannot predict every situation in life and lawful action is not always the most morally appropriate one''. 

Then we can attack argument a (which advocates for continuing forward). This will be yet another deontological argument, which we will mark it by j and define as: ``ABS systems are installed in almost all traditional vehicles in order to make the driver safer. The system is in place, although this could endanger other participants in traffic and pedestrians, as the car needs to travel a longer distance before coming to a stop. Therefore - it is not immoral to prioritise the driver''. It is important to note that ABS in this context stands for Anti-lock Braking System, which is an automobile safety system that generally offers improved vehicle control and decreases stopping distances on dry and slippery surfaces; however, on loose gravel or snow-covered surfaces, ABS can significantly increase braking distance, although still improving vehicle steering control \footnote{More information on ABS, its benefits and potential disadvantages available at
\url{https://www.scienceabc.com/innovation/abs-sensors-anti-lock-breaking-system-technology-cars-work.html} (last visited: 2018-02-28)}.

Now we will add another deontological argument (k), but this time we will attack argument b (which argues for changing the lane in order to save a life). Argument k is: ``It would be morally unacceptable to kidnap a healthy person and harvest their organs even if that action would save two lives. For the same moral reasons, it is immoral to change the lane and kill an innocent driver''. 

One could claim however that both j and k are taking the analogues too far, which will be argument l: ``The moral dilemma of autonomous car ethics is unique due to previously incomprehensible technological capabilities that are inherent to the the self-driving cars. Therefore, moral decisions from the past are not appropriate guidelines". This argument could, of course be attacked by an argument m, which is consequentialist in nature: ``Being indecisive and avoiding to learn from the situations in the past can lead to greater harm in the future''.

Now, let us attack j by two more arguments, one in D and another in C. The deontological argument is ``The purpose is of the ABS is simply to prevent the driver from braking in a way that would be counterproductive, so saying that the intent including ABS in traditional cars to prioritise the driver over the other participants in the traffic is a misinterpretation'' (n). The consequentialist argument is: ``ABS is not the ideal approach to copy, because ABS can actually cause more harm in certain situations (for example when trying to stop on an icy road in the moment after the driver has seen that there is a hole in the road)'' (o).

Finally, let us add two more arguments that will be factual. We do this to show that Bench-Capon \cite{capon02, dunne2002coherence, bench2003persuasion} represents facts in the VAF as arguments that are always to be prioritised, regardless of value system. The first argument p attacks both a and b by saying: ``The car would avoid any direct impact if it would crash into one of the walls of the tunnel''. This argument could be easily dismissed by another factual argument q which would state that ``there is no difference between choosing the lane to stay in and choosing the wall to crash into, as in the both cases the consequences and the moral implications of the decision remain unchanged''. 

\begin{figure}[p]
    \centering
    \fbox{
    \begin{minipage}{0.95\textwidth}
        \includegraphics[width=14cm]{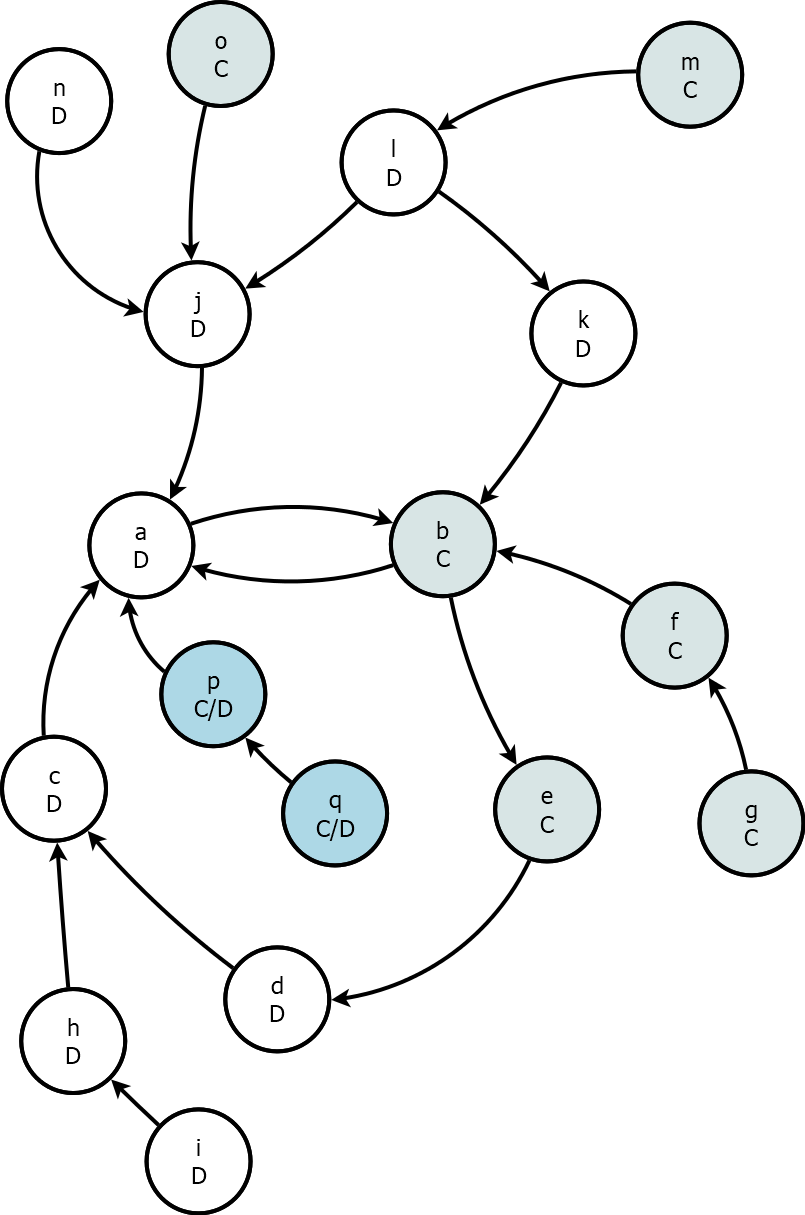}
    \end{minipage}
    }
    \caption{Value-based argumentation framework for the expended problem}
    \label{mdgraph2}
\end{figure}

The graphical representation of this argumentation framework is given in figure \ref{mdgraph2}.

Now let us analyse the graph in figure \ref{mdgraph2}. Firstly, we can conclude that the status of arguments g, f, e and d will not change.

We no longer have a chain dca, since c is now attacked by multiple arguments (definition \ref{vafchainNEWdef}). This means that c will be the last argument of chains dc and ihc, and the parity of c is even, as it is even in chain dc - as in the first rule of property \ref{vafmultiplechainsprop}. Therefore, c is still indefensible according to the path 5. This is the case although argument h, which also attacks c, is defeated by i. It is clear that h is indefensible according to the path 5, whereas i is on path 1 and thus objectively acceptable.

Arguments m, o, and n are not attacked by any argument and are therefore on path 1 and objectively acceptable. Argument l is subjectively acceptable, as it is on path 2. We could further note that argument l follows the second rule of property \ref{vafmultiplechainsprop} and is thus the first argument of chains lj and lk. We can also note that j is the last argument of both nj and lj and thus follows the first rule of property \ref{vafmultiplechainsprop}. Since j is attacked by multiple arguments, j and a are not in the same chain (definition \ref{vafchainNEWdef}). We can also note that j is indefensible, as it is on path 4, since it is directly attacked by odd chains of both values.  On the other hand, argument k is subjectively acceptable, since it is on path 3. 

Arguments p and q represent facts. This means that their value is above both C and D. However, facts can still defeat facts (definition \ref{vafdefeatsdef}). Therefore we can say that q is objectively acceptable and p is indefensible.

We have seen that j, c, and p are indefensible. Therefore, the only argument that could influence a is b (or, according to our approach in section \ref{vafproperties} - the chain gfb). Since k is also indefensible, both a and b will remain subjectively acceptable.

So to summarise, we can classify the arguments in the following way now:
\begin{itemize}
    \item Objectively acceptable: d (D), g (C), i (D), m (C), n (D), o (C), q (C/D);
    \item Acceptable iff C \(>\) D: b (C), k (D);
    \item Acceptable iff D \(>\) C: a (D), e (C), l (D);
    \item Indefensible: c (D), f (C), h (D), j (D), p (C/D).
\end{itemize}

The preferred extension in case when deontology is preferred is \{a, d, e, g, i, k, m, n, o, q\}. In case when consequentialism is preferred, the preferred extension is \{b, d, g, i, k, m, n, o, q\}. In addition to conclusions from section \ref{mdproblemdescription}, the rational agent would also have to accept (regardless of its preferred ethical approach) that: sometimes it is moral not to follow the law (i); the analogy with ABS is not an optimal analogy (o, n); and that it is pointless to turn go off road (q).

\section{\change{Attempting to} \vthree{Formalise the Claims in Propositional Logic}}

\change{In this section, an attempt \vthree{to formalise the claims of arguments in} propositional \vthree{logic} will be presented. The failings of this attempt will showcase that the propositional logic is not expressive enough for modelling of the moral dilemma, which is an insurmountable challenge, because i}n order to be able to analyse the problem from the point of view offered by semi-abstract argumentation framework (SAF) and semi-abstract value-based argumentation framework (SVAF), it is necessary to translate the arguments from natural language into the statements of predicate logic. 

Being a logic without quantification \change{and modality operators}, knowledge representation in propositional logic is limited\change{, as further elaborated in} \cite{van2008handbook}. \change{Therefore it is immediately clear that} some of the statements will have to be adapted from their original form presented in section \ref{mdproblemdescription}. Let us start \change{the proposed ``na\"ive'' modelling} with listing the propositional variables which will be used. For variable names two unabbreviated words connected with an underscore and single unabbreviated words will be used. On one hand, this naming enables easier reading of the indented meaning behind the formulae containing the variables. On the other hand, this will helps in differentiating the variables (full words) from the argument labels (lower case single letters) and values (capitalised single letters). For the modelling of the simpler version of the problem (c.f. section \ref{mdproblemdescription}), nine propositional variables are necessary. 

\begin{enumerate}
    \item $murder$ - the car murders people
    \item $stay$ - the car stays in this (right) lane
    \item $fewer$ - the car kills fewer people
    \item $switch$ - the car switches to the other (left) lane
    \item $insider\_dies$ - the driver of the autonomous car dies
    \item $kill$ - people are killed
    \item $prioritise\_insider$ - the car prioritises saving the driver
    \item $buy\_autonomous$ - it is important that the people are willing to buy autonomous cars
    %\item $save\_driver$ - the driver survives
    \item $endanger$ - other participants in the traffic are are endangered by the intoxicated driver
\end{enumerate}

The translations of the arguments is given in table \ref{tabTranslations}. 

%\begin{table}[h]
%\centering
%\caption{Translations of the arguments into propositional formulae}
%\label{tabTranslations}
\begin{longtable}{|p{.06\textwidth}|p{.04\textwidth}|p{.30\textwidth}|p{.25\textwidth}|p{.20\textwidth}|}
\hline
\textbf{\change{Label}} & \textbf{Val.} & \textbf{Statement} & \textbf{Formula (claim)} & \textbf{Note} \\ \hline
equi & fact &  & 
$(stay \land \neg switch)\ \lor\ (\neg stay \land switch)$ & Staying means not switching and vice versa. \\ \hline
a & D & The car should \change{never take any actions that results in the death of} traffic participants. & 
$(stay \supset \neg murder) \ \land \ (switch \supset murder)$ & Switching to the other lane would kill other participants. \\ \hline
b & C & The car should change the lanes if that leads to fewer dead people. & 
$(switch \supset fewer) \ \land \ (stay \supset \neg fewer)$ & Switching to the other lane will kill fewer people. \\ \hline
c & D & Killing the driver (of the autonomous car) leads to a death too. Furthermore, staying in the lane would kill the driver (of the autonomous car).& 
$(insider\_dies \supset kill) \ \land \ (stay \supset insider\_dies)$ &  \\ \hline
d & D & Letting someone die is not a murder. & 
$kill$ & \\ \hline
e & C & Not prioritising the driver within will make fewer people buy autonomous vehicles, thus prioritise the driver. Furthermore, prioritising the driver of the autonomous car means that they should be saved (if possible). & 
$(buy\_autonomous \supset prioritise\_insider) \ \land \ (prioritise\_insider \supset switch)$ &  In this situation, switching lanes means letting the driver car live. \\ \hline
f & C & Switching lanes means allowing the intoxicated driver to continue on, endangering others. & 
$switch \supset endanger$ &  \\ \hline
g & C & Potential harm is better than a certain additional death. & 
$endanger$ &  \\ \hline
\caption{\vthree{Formalisation of claims of arguments}}
\label{tabTranslations}
\end{longtable}
%\end{table}

Let us now consider attacks in this framework (cf. figure \ref{mdgraph1}).

\begin{align}
a \rightarrow b:\ & (stay \supset \neg murder) \ \land \ (switch \supset murder) \rightarrow (switch \supset fewer) \ \land \ (stay \supset \neg fewer) \notag\\
b \rightarrow a:\ & (switch \supset fewer) \ \land \ (stay \supset \neg fewer) \rightarrow (stay \supset \neg murder) \ \land \ (switch \supset murder) \notag\\
c \rightarrow a:\ & (insider\_dies \supset kill) \ \land \ (stay \supset insider\_dies) \rightarrow  \notag\\& (stay \supset \neg murder) \ \land \ (switch \supset murder) \notag\\
d \rightarrow c:\ & kill \rightarrow (insider\_dies \supset kill)\ \land\ (stay \supset insider\_dies) \notag\\
e \rightarrow d:\ & (buy\_autonomous \supset prioritise\_insider) \ \land \ (prioritise\_insider \supset switch) \rightarrow kill  \notag\\
b \rightarrow e:\ & (switch \supset fewer) \ \land \ (stay \supset \neg fewer)  \rightarrow \notag\\
&(buy\_autonomous \supset prioritise\_insider) \ \land \ (prioritise\_insider \supset switch) \notag\\
f \rightarrow b:\ & switch \supset endanger \rightarrow (switch \supset fewer) \ \land \ (stay \supset \neg fewer) \notag\\
g \rightarrow f:\ & endanger \rightarrow  switch \supset endanger \notag
\end{align} 

\change{At this point, we have to note the reasons why this translation indeed is unsatisfactory.}

\change{\begin{itemize}
\item In order to model the auxiliary verb \emph{should}, we require deontoic logic \vthree{from \cite{von1951deontic}}.
    \item Lack of quantifiers is very problematic (for example in the case of $kill$).
    \item The meaning of individual variables inescapably varies (compare the meaning of $kill$ in c and d).
    \item The translations from table \ref{tabTranslations} deviate from the original formulations, making the decision to leave attack relations unaltered problematic.
\end{itemize}}

\change{One could avoid some of the issues by having variables that are more closely related to the arguments (e.g. having the whole claim of an argument be represented by a single variable). The problem then would be that no additional attacks could be inferred using attack principles. Thus the SAF and the SVAF would offer no advantages over the analyses over Dung's framework and the VAF, respectively.}

\change{Therefore it must be concluded that the propositional logic is not expressive enough for reasoning about this ethical dilemma. Therefore, the SAF and the SVAF (as introduced in sections \ref{safconcepts} and \ref{myArgSection}) cannot be applied without prior adaptation of the way in which the claims of the arguments are formalised. Such an adaptation would have to incorporate more expressive logics.}

\chapter{Summary and Future Work}

This work presented and discussed concepts from Dung's \cite{dung95} argumentation framework, Bench-Capon's \cite{capon02, bench2003persuasion} value-based argumentation framework, and Ferm\"uller\change{'s} and Corsi's \cite{corfer17lori, pfefer18} semi-abstract argumentation framework. The paper then proposed a new framework which unites the last two. \change{Dung's and Bench-Capon's} frameworks were then used in an analysis of a moral dilemma concerning the ethics of autonomous vehicles. The Dung's framework enabled a basic analysis of the problem. By using value-based framework this analysis was extended and incorporated objectively and subjectively acceptable arguments presented by the moral dilemma. In order to apply the semi-abstract \change{(value-based)} framework, the original dilemma \change{would have} to be reformed into a set of propositional formulae. \change{However, such a transformation was found to be infeasible due to too low expressivity of propositional logic.}

Further research is necessary in order to discover how the newly-introduced semi-abstract value-based argumentation relates to the described argumentation chain characteristics from \cite{capon02} (c.f. property \ref{vafmultiplechainsprop}) and if there is a way to relate this framework to traditional logics, as was done with SAF in \cite{pfefer18}. Likewise, it might be significant to expand the newly-proposed framework to incorporate probabilistic and fuzzy logics, as was done with semi-based argumentation in \cite{corfer17lori} and \cite{corfer18fuzzyArgumentation}.

Furthermore, the example of uniting these two (at first sight) thoroughly different frameworks raises the question of whether value-based or semi-abstract frameworks could also be combined with or extended by other frameworks such as the ones based on contextual preferences \cite{amgoud2000argumentation, modgil2009reasoning}, sceptical assumption-based argumentation \cite{dung2006dialectic}, persuasion \cite{bench2002agreeing, dunne2003two}, or dialectics \cite{Brewka2017AbstractDF}. Moreover, each of the possible combinations would have to provide answers on questions regarding the logical fondness and possible applications.

\backmatter

% Use an optional list of figures.
\listoffigures % Starred version, i.e., \listoffigures*, removes the toc entry.

% Use an optional list of tables.
\cleardoublepage % Start list of tables on the next empty right hand page.
\listoftables % Starred version, i.e., \listoftables*, removes the toc entry.

% Use an optional list of alogrithms.
% \listofalgorithms
% \addcontentsline{toc}{chapter}{List of Algorithms}

% Add an index.
\printindex

% Add a glossary.
\printglossaries

% Add a bibliography.
\bibliographystyle{alpha}
\bibliography{thesis}

\end{document}